\documentclass{article}

\usepackage[accepted]{icml2025}

\usepackage{amsmath}
\usepackage{amssymb}
\usepackage{mathtools}
\usepackage{amsthm}

\icmltitlerunning{SparseLoRA: Accelerating LLM Fine-Tuning with Contextual Sparsity}

\usepackage{color,xcolor}
\usepackage{epsfig}
\usepackage{graphicx}

\usepackage{adjustbox}
\usepackage{array}
\usepackage{booktabs}
\usepackage{colortbl}
\usepackage{float,wrapfig}
\usepackage{hhline}
\usepackage{multirow}
\usepackage{subcaption} 
\usepackage[toc,page]{appendix}
\usepackage{makecell}

\usepackage{amsmath,amsfonts,amsthm,amssymb}
\usepackage{bm}
\usepackage{nicefrac}
\usepackage{microtype}
\usepackage{inconsolata}
\usepackage{pifont}

\usepackage{changepage}
\usepackage{extramarks}
\usepackage{fancyhdr}
\usepackage{lastpage}
\usepackage{setspace}
\usepackage{soul}
\usepackage{xspace}
\usepackage{indentfirst}
\usepackage{paralist}
\usepackage{booktabs,arydshln}

\usepackage[pagebackref=true,breaklinks=true,letterpaper=true,colorlinks,citecolor=darkblue,bookmarks=false]{hyperref}
\usepackage{url}

\usepackage{algorithm, algorithmic}
\usepackage{enumerate}
\usepackage{lipsum}
\usepackage[frozencache,cachedir=minted]{minted}
\newcolumntype{L}[1]{>{\raggedright\let\newline\\\arraybackslash\hspace{0pt}}m{#1}}
\newcolumntype{C}[1]{>{\centering\let\newline\\\arraybackslash\hspace{0pt}}m{#1}}
\newcolumntype{R}[1]{>{\raggedleft\let\newline\\\arraybackslash\hspace{0pt}}m{#1}}

\newcommand{\fig}[1]{Figure~\ref{#1}}

\newcommand{\ignorethis}[1]{}

\makeatletter
\DeclareRobustCommand\onedot{\futurelet\@let@token\@onedot}
\def\@onedot{\ifx\@let@token.\else.\null\fi\xspace}

\def\eg{\emph{e.g}\onedot}

\makeatother

\makeatletter
\def\adl@drawiv#1#2#3{%
        \hskip.5\tabcolsep
        \xleaders#3{#2.5\@tempdimb #1{1}#2.5\@tempdimb}%
                #2\z@ plus1fil minus1fil\relax
        \hskip.5\tabcolsep}
\newcommand{\cdashlinelr}[1]{%
  \noalign{\vskip\aboverulesep
           \global\let\@dashdrawstore\adl@draw
           \global\let\adl@draw\adl@drawiv}
  \cdashline{#1}
  \noalign{\global\let\adl@draw\@dashdrawstore
           \vskip\belowrulesep}}
\makeatother

\newcommand{\xmark}{\ding{55}}

\definecolor{citecolor}{HTML}{0071bc}
\definecolor{mydarkblue}{rgb}{0,0.08,1}
\definecolor{mydarkgreen}{rgb}{0.02,0.6,0.02}
\definecolor{mydarkred}{rgb}{0.8,0.02,0.02}
\definecolor{mydarkorange}{rgb}{0.40,0.2,0.02}
\definecolor{mypurple}{RGB}{111,0,255}
\definecolor{myred}{rgb}{1.0,0.0,0.0}
\definecolor{mygold}{rgb}{0.75,0.6,0.12}
\definecolor{mydarkgray}{rgb}{0.66, 0.66, 0.66}

\definecolor{darkblue}{rgb}{0,0.08,1}
\definecolor{darkgreen}{rgb}{0.02,0.6,0.02}
\definecolor{darkred}{rgb}{0.8,0.02,0.02}
\definecolor{darkorange}{rgb}{0.40,0.2,0.02}
\definecolor{darkpurple}{RGB}{111,0,255}

\definecolor{spc}{RGB}{119, 107, 170}
\definecolor{pct}{rgb}{0.7, 0, 0.2}

\newcommand{\ssect}[1]{\S~\ref{#1}}
\newcommand{\append}[1]{Appendix~\ref{#1}}
\newcommand{\tbl}[1]{Table~\ref{#1}}
\newcommand{\algo}[1]{Algorithm~\ref{#1}}

\newcommand{\myparagraph}[1]{\vspace{0pt}\paragraph{#1}}

\definecolor{mydarkblue}{rgb}{0,0.08,1}

\def\method{SparseLoRA\xspace}

\begin{document}

\twocolumn[
\icmltitle{SparseLoRA: Accelerating LLM Fine-Tuning with Contextual Sparsity}

\icmlsetsymbol{equal}{*}
\icmlsetsymbol{lead}{$\dagger$}

\begin{icmlauthorlist}
\icmlauthor{Samir Khaki}{equal,ut}
\icmlauthor{Xiuyu Li}{equal,lead,ucb}
\icmlauthor{Junxian Guo}{equal,mit}
\icmlauthor{Ligeng Zhu}{mit}
\icmlauthor{Chenfeng Xu}{ucb}
\icmlauthor{Konstantinos N. Plataniotis}{ut}
\icmlauthor{Amir Yazdanbakhsh}{gdm}
\icmlauthor{Kurt Keutzer}{ucb}
\icmlauthor{Song Han}{mit}
\icmlauthor{Zhijian Liu}{mit}
\end{icmlauthorlist}

\centering\url{https://z-lab.ai/projects/sparselora}

\icmlaffiliation{ut}{University of Toronto}
\icmlaffiliation{ucb}{UC Berkeley}
\icmlaffiliation{mit}{MIT}
\icmlaffiliation{gdm}{Google DeepMind}

\icmlcorrespondingauthor{Samir Khaki}{samir.khaki@mail.utoronto.ca}

\vskip 0.3in
]

\printAffiliationsAndNotice{\icmlEqualContribution \textsuperscript{$\dagger$}Project lead}

\begin{abstract}

Fine-tuning LLMs is both computationally and memory-intensive. 
While parameter-efficient fine-tuning methods, such as QLoRA and DoRA, reduce the number of trainable parameters and lower memory usage, they do not decrease computational cost.
In some cases, they may even slow down fine-tuning.
In this paper, we introduce \textit{\method}, a method that accelerates LLM fine-tuning through contextual sparsity.
We propose a lightweight, training-free \textit{SVD sparsity estimator} that dynamically selects a sparse subset of weights for loss and gradient computation.
Also, we systematically analyze and address sensitivity across layers, tokens, and training steps.
Our experimental results show that \method reduces computational cost by up to \textbf{2.2$\times$} and a measured speedup of up to \textbf{1.6$\times$} while maintaining accuracy across various downstream tasks, including commonsense and arithmetic reasoning, code generation, and instruction following.
\end{abstract}
\section{Introduction}

Large language models (LLMs) are trained on vast, general-domain datasets. They are often fine-tuned to improve their performance in specific domains~\cite{saab2024medgemini} or to align their predictions with user preferences~\cite{zhang2024llama}. However, fine-tuning very large models can be prohibitively expensive, both in terms of memory requirements and computational costs.

\begin{figure}[t]
    \centering
    \includegraphics[width=\linewidth]{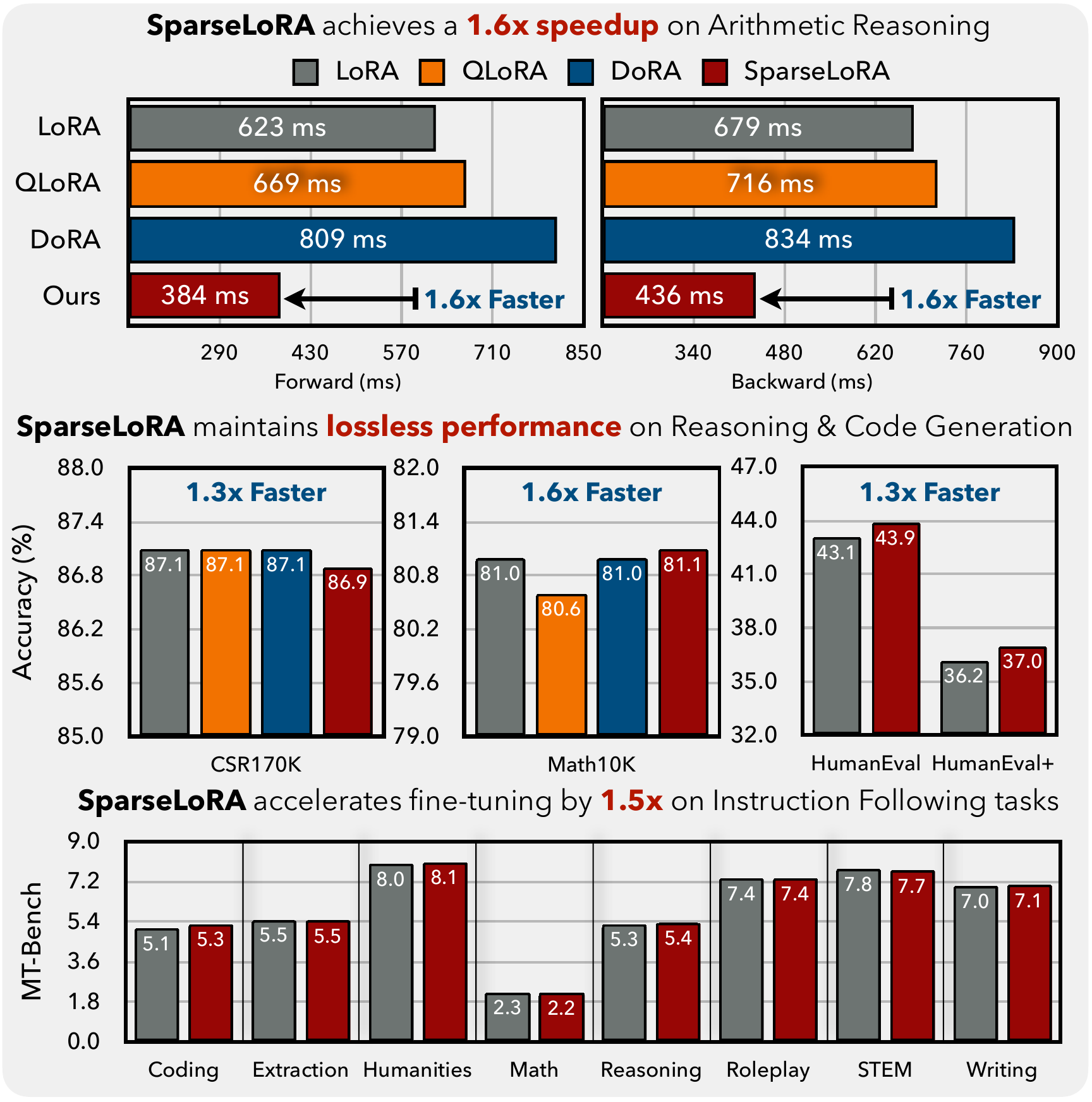}
    \caption{Many recent parameter-efficient fine-tuning methods, such as QLoRA and DoRA, do not reduce compute. Our~\method accelerates LLM fine-tuning with minimal accuracy loss across a range of downstream tasks, including commonsense reasoning, math reasoning, code generation, and instruction following. See Section~\ref{sec:experiments} for more details.}
    \label{fig:teaser}
    \vspace{-5mm}
\end{figure}

Extensive efforts have been made in parameter-efficient fine-tuning (PEFT) to reduce memory consumption of LLM fine-tuning. LoRA~\citep{lora} represents weight updates using low-rank approximations. Building upon this, many follow-up methods~\citep{dettmers2023qlora,dora} have been proposed to further reduce the number of trainable parameters. While they are effective in reducing memory usage, they do not reduce computation. In fact, they can sometimes slow down fine-tuning due to the overhead they introduce: DoRA is 20\% slower than LoRA (see \fig{fig:teaser}).

In this paper, we present \textbf{SparseLoRA} to accelerate LLM fine-tuning with contextual sparsity, making it both memory- and computation-efficient.
Contextual sparsity has already been used in accelerating LLM inference~\citep{dejavu}. 
\method shows for the first time that it can also play a role in LLM fine-tuning, where (1) only a sparse subset of weights is required for loss and gradient computation, and (2) this sparse subset needs to be determined based on the input sequence or tokens. 
To realize this, we propose using an \textbf{SVD sparsity estimator} to identify which channels should be activated based on samples within each batch. 
It is very \textit{lightweight}, adding only 0.05\% FLOPs and 0.8\% runtime overhead to fine-tuning. It is \textit{training-free}, unlike the look-ahead predictor proposed in Deja Vu~\cite{dejavu}, which leads to better generalization across datasets.

We have conducted systematic sensitivity analysis across multiple dimensions. 
First, each layer responds differently to sparsity, so we apply non-uniform sparsity based on layer sensitivity analysis. 
Second, output tokens are much more sensitive to pruning than context tokens, so we apply sparsity only to context tokens. 
Finally, early iterations in fine-tuning are more sensitive, so we run dense fine-tuning in the early iterations and switch to sparse fine-tuning for the remainder.

\begin{figure}[!t]
    \centering
    \includegraphics[width=\linewidth]{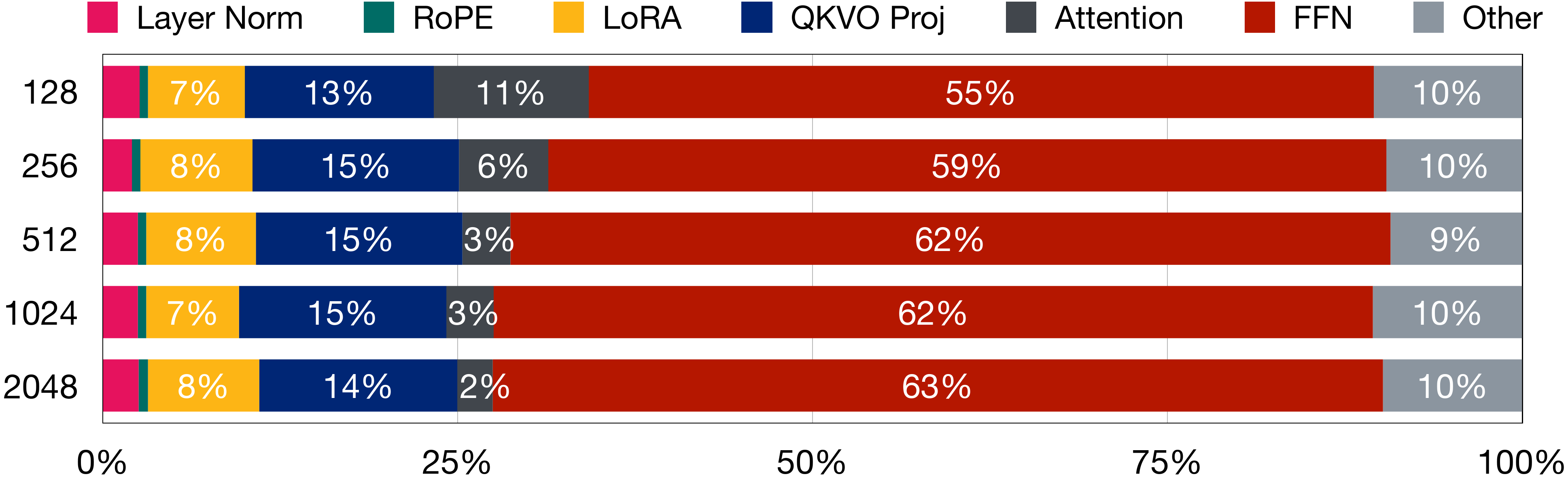}
    \caption{Runtime breakdown of LLaMA3-8B fine-tuning under different sequence lengths.}
    \label{fig:lora-profile}
    \vspace{-5mm}
\end{figure}

Evaluated across a diverse set of benchmarks, \method achieves a computational cost reduction of up to 2.2$\times$ and a wall-clock speedup of up to 1.6$\times$ while maintaining accuracy on various downstream tasks, including commonsense and arithmetic reasoning, code-generation, and complex instruction following. 
To the best of our knowledge, this is the first work to leverage contextual sparsity for accelerating LLM fine-tuning. 
We believe that our work will inspire future research into fine-tuning methods that optimize both parameter and computational efficiency.
%
%
\begin{figure*}[!t]
    \centering\includegraphics[width=0.9\textwidth]{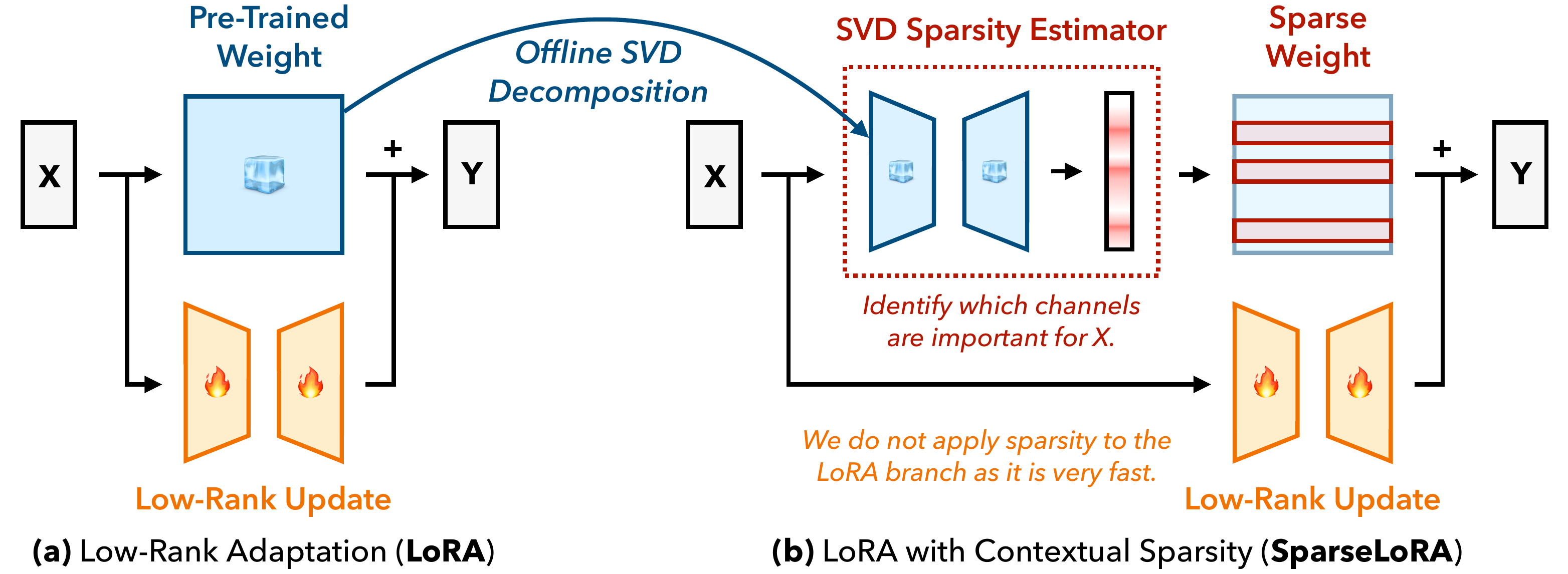}
    \caption{SparseLoRA accelerates LLM fine-tuning with contextual sparsity. It first performs an offline SVD decomposition of the pre-trained weights to construct the SVD sparsity estimator. During fine-tuning, it relies on the SVD sparsity estimator to identify which weight channels are needed for the given input sequence. It then sparsely computes by slicing the weights on the fly. Note that sparsity is applied only to the main branch, as the LoRA branch is typically very fast.}
    \label{fig:overview}
\end{figure*}
\section{Related Work}

\myparagraph{Contextual Sparsity in LLMs.}

Sparsity has long been employed to enhance the efficiency of neural networks~\citep{han2015deep_compression, han2015learning, han2016eie, frantar2023sparsegpt, sun2024wanda}. Recently, there has been research on a more dynamic approach: accelerating LLMs inference by leveraging contextual sparsity~\citep{dejavu, song2023powerinfer, xue2024powerinfer2, alizadeh2024flash, lee2024cats, akhauri2024shadowllm, liu2024trainingfree} at test time. Unlike static sparsity, contextual sparsity is a phenomenon where significant portions of a model's hidden states dynamically contain zero-valued neurons based on the input context. This allows for input-dependent sparse computation without compromising outcomes. While such activation sparsity naturally emerges in ReLU-based FFNs~\citep{li2023lazyneurons, mirzadeh2023relu, dejavu, alizadeh2024flash}, newer architectures often employ non-ReLU activations~\citep{touvron2023llama2, dubey2024llama3, team2024gemma, team2024gemma2} that create different sparsity patterns, necessitating alternative methods to reintroduce and exploit sparsity. 

Recent work has explored various techniques, including continued pretraining~\citep{song2023powerinfer, zheng2024learn, zhang2024relu2, song2024prosparse, song2024turbo, xue2024powerinfer2} and magnitude pruning with specific metrics~\citep{lee2024cats, akhauri2024shadowllm, liu2024trainingfree}, to leverage sparsity in LLMs. These approaches aim to reduce memory usage and computation time, with some work deploying small neural networks to predict non-zero activations~\citep{dejavu, alizadeh2024flash, akhauri2024shadowllm, song2023powerinfer, xue2024powerinfer2}. In this work, we propose a novel approach to extend the benefits of contextual sparsity to the fine-tuning process for the first time, which addresses the associated challenges and accelerates fine-tuning without compromising performance.

\myparagraph{Memory-Efficient Fine-tuning.}

As language models grow larger, memory-efficient fine-tuning methods have become crucial. Parameter-efficient fine-tuning (PEFT) techniques address this challenge by updating only a small subset of parameters. LoRA~\citep{lora} employs low-rank matrices to adjust pretrained model weights, sparking a rich line of research with numerous works proposing improvements and variations~\citep{dettmers2023qlora, shi2023toast, qiu2023oft, longlora, kopiczko2024vera, dora, meng2024pissa, hayou2024loraplus, wang2024loraga, wang2024lorapro, liu2024butterflyoft, pan2024lisa, yang2024sft}. Among these, DoRA~\citep{dora} reparameterizes weight matrices to achieve more effective optimization. QLoRA~\citep{dettmers2023qlora} combines quantization with low-rank adapters, enabling fine-tuning of large models on a single GPU. 

Recent research has also focused on exploiting the low-rank structure of weight gradients. GaLore~\citep{zhao2024galore} and its weight-quantized variant~\citep{zhang2024qgalore} leverage this property to reduce optimizer state memory, while WeLore~\citep{jaiswal2024welore} investigates how low-rank weights emerge from low-rank gradients during training. While these advancements make LLM adaptation more memory-efficient and accessible, they primarily focus on reducing memory usage, sometimes even at the cost of increased computation time. Our approach addresses this missing piece by focusing on compute efficiency, complementing existing memory-efficient techniques to enable truly resource-efficient fine-tuning.

\myparagraph{Computation-Efficient Training.}

Prior work has explored sparsity to accelerate LLM training through various approaches~\citep{spdf, longlora, mozaffari2024slope}. LongLoRA~\citep{longlora} enables efficient context window extension via shifted sparse attention during fine-tuning while retaining standard attention at inference, reducing quadratic computation costs. General sparse training techniques have demonstrated promising FLOPs reduction through combinations of sparse pre-training and further fine-tuning~\citep{spdf, mozaffari2024slope}. Other methods employ dynamic sparsity patterns through gradient-based selection~\citep{zhou2021efficient, liijcai2022p586}. Yet these techniques either focus on long-context scenarios where attention dominates computation, or rely on unstructured sparsity, whose theoretical gains are challenging to realize on consumer-grade GPUs. While there have been recent attempts to apply structural sparsity~\citep{ma2024sparsityaccelerated, chen2025celora}, they either are not memory efficient or accelerate only partial fine-tuning computation (e.g., the backward pass), limiting their effectiveness. Our work introduces structured contextual sparsity with training-free dynamic selection, enabling practical end-to-end acceleration while remaining memory-efficient.

\section{Method}

In this section, we present the design of our \method. As shown in Figure~\ref{fig:lora-profile}, linear layers dominate LoRA fine-tuning runtime in conventional settings. Therefore, we apply \textbf{dynamic channel sparsity} to the main branch of LoRA fine-tuning while keeping the LoRA branch dense. This approach selectively activates only the most important neurons in FFN and attention layers. Since the main branch accounts for the vast majority of computation, and the sparsity introduced is structural and hardware-friendly, we achieve significant efficiency improvements without altering which parameters are updated. By sparsifying only the base model while keeping LoRA intact, \method maintains both \textbf{memory- and computation-efficient} fine-tuning with no impact on inference performance.

\subsection{Sparse Neuron Selection Criteria}

To achieve compute-efficient fine-tuning while maintaining effectiveness, we require dynamic rather than static sparsity patterns that adapt to each input. 
Prior research has explored contextual activation sparsity in LLM inference~\citep{dejavu, alizadeh2024flash, akhauri2024shadowllm, liu2024trainingfree}. However, these \textit{inference} methods target single-token computations in auto-regressive generation and do not directly translate to \textit{fine-tuning}, which has distinct workloads that consist of multiple sequences of tokens in a batch. To bridge this gap, we first define \textbf{“oracle”} criteria for fine-tuning neuron selection using intermediate activations, establishing an ideal but computationally infeasible metric. This oracle then guides the practical development of efficient approximations that enable on-the-fly channel selection for sparse computation. We categorize the linear layers in LLMs into three types -- FFN, VO projections, and QK projections -- and propose two oracle criteria tailored to their properties as follows.

\begin{figure*}[htbp]
    \centering
    \includegraphics[width=\textwidth]{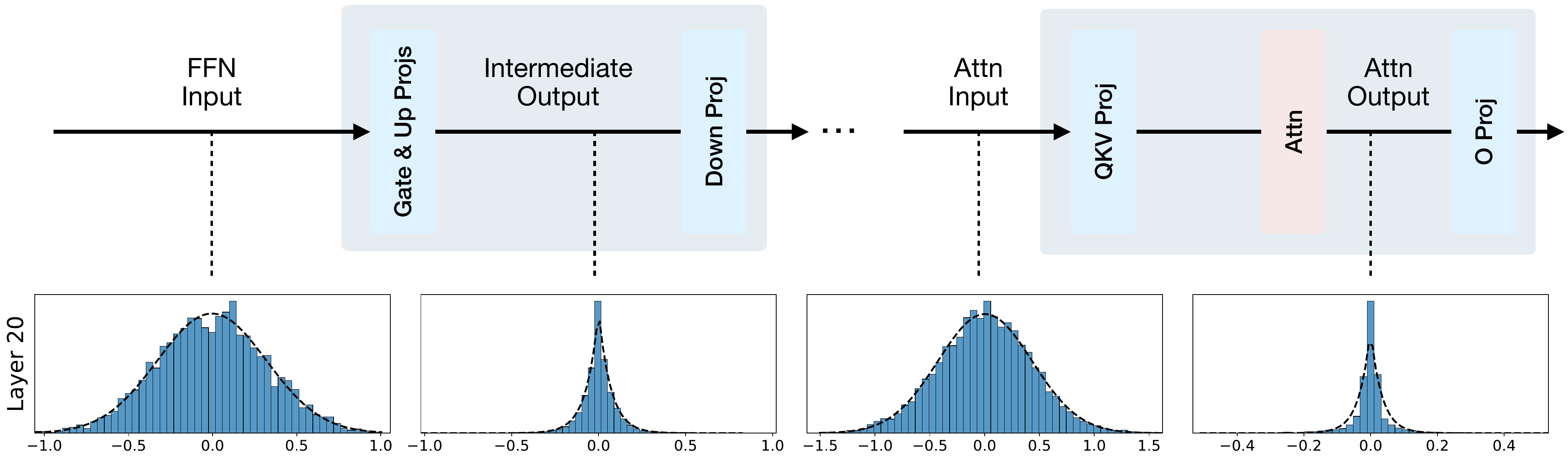}
    \caption{Activation value distributions across all tokens of a sequence for layer 20 of LLaMA2-7B. Each column shows inputs to: $W_{gate,up}$, $W_{down}$ of FFNs, and $W_{Q,K,V}$, $W_{O}$ of attention projections. The inputs to $W_{down}$ and $W_{O}$ follow Laplace distributions, enabling higher sparsity when using the L2 norm metric. $W_{gate,up}$ is pruned alongside corresponding $W_{down}$ channels and $W_V$ is pruned alongside corresponding $W_O$ channels, while $W_{Q,K}$ requires a different criterion.}
    \label{fig:act_distributions}
\end{figure*}



\subsubsection{Selection with L2 Norm}
\label{ssect:ffn-oracle}

\begin{figure}[t]
    \centering
    \includegraphics[width=\linewidth]{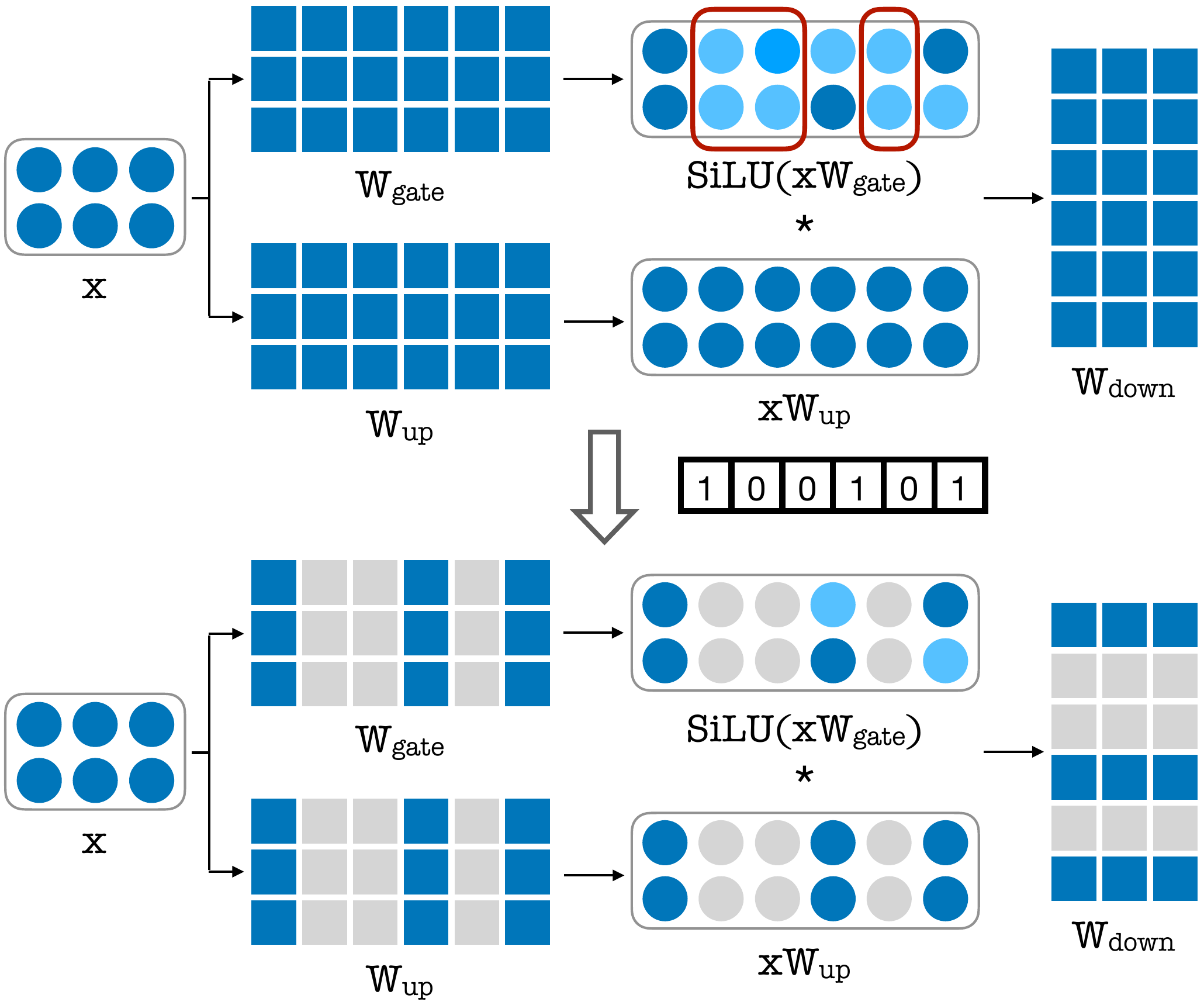}
    \caption{An illustration of how sparsity is applied to FFNs.}
    \label{fig:ffn-oracle}
\end{figure}

Motivated by the extreme sparsity in input activations for certain linear layers, such as the SiLU-induced sparsity in the down projection of FFNs~\cite{alizadeh2024flash, lee2024cats, song2023powerinfer} as shown in~\fig{fig:act_distributions}, we introduce an L2 Norm metric to identify and retain the most significant neurons as the first oracle criterion. Importantly, this approach allows us to naturally extend the sparsity pattern to the preceding linear layers -- specifically using FFN as an example, the channel indices selected for the down projection can be applied to both up and gate projections, as illustrated in~\fig{fig:ffn-oracle}.

While prior work has applied similar magnitude-based metrics for inference with single-token~\citep{dejavu, lee2024cats, akhauri2024shadowllm, liu2024trainingfree}, we extend this to fine-tuning by considering cumulative activations across all samples and tokens in a batch. Despite recent exploration of incorporating gradient and weight information~\citep{akhauri2024shadowllm, sun2024wanda}, we find the L2 norm of activations alone provides a simple yet effective criterion, as activations generally contain more influential outliers than weights~\citep{xiao2023smoothquant}. We apply this metric to both FFNs and the outer channels of value and output (VO) projections in attention, where input activations to output projections show similar sparsity characteristics as demonstrated in~\fig{fig:act_distributions}.





\subsubsection{Selection with QK Norm}
For attention blocks, we aim to sparsify the linear layers as well: query, key, value, and output projections. While the L2 Norm metric works well for VO projections, it proves impractical for QK projections due to their lower input activation sparsity as shown in~\fig{fig:act_distributions}.
Previous approaches exploring contextual sparsity during inference have proposed identifying and pruning unimportant attention heads~\citep{dejavu, akhauri2024shadowllm}. However, this head-level pruning strategy proves problematic for fine-tuning scenarios that process multiple tokens simultaneously. Unlike FFNs where we can selectively choose from 11,384 channels in LLaMA2-7B, pruning entire attention heads (e.g., removing 1 out of 32) significantly constrains our pruning granularity and risks losing critical information. We empirically verify in~\ssect{ssect:ablation} that this coarse-grained approach leads to degraded performance. Additionally, a detailed analysis of attention head activation patterns during fine-tuning is provided in~\append{append:attn_head}.

To address this challenge, we introduce an oracle criterion for sparsifying QK projections based on attention scores, targeting channels with minimal contributions. Specifically, we define a proxy metric that quantifies each channel’s importance and sparsify those with the lowest values. Given query and key projections \( \mathbf{Q}, \mathbf{K} \in \mathbb{R}^{(B \times L) \times D} \), where \( B \) is the batch size, \( L \) is the sequence length, and \( D \) is the hidden dimension, we compute their L2 norms across the flattened batch and sequence dimensions:  
\[
\mathbf{q} = \|\mathbf{Q}\|_2, \quad \mathbf{k} = \|\mathbf{K}\|_2.
\]
The element-wise product of these normalized scores serves as our importance metric:  
\[
\mathbf{s} = \mathbf{q} \odot \mathbf{k}.
\]
We retain the top \( n \) channels based on \( \mathbf{s} \), where \( n \) is determined by the desired sparsity ratio.

\begin{figure*}[t]
    \centering
    \includegraphics[width=\textwidth]{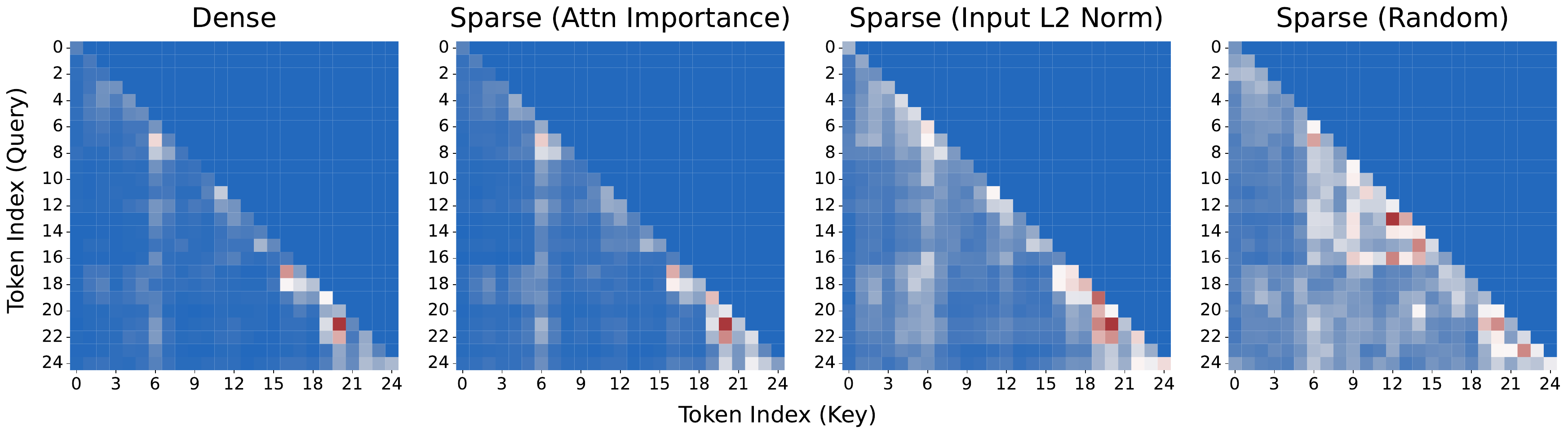}
    \caption{Comparison of attention maps for different pruning strategies at 50\% sparsity. The figure shows the attention maps for the last 25 tokens of head 7 using dense full attention, pruning based on attention importance (ours), pruning based on input L2 norm (simlar to FFNs), pruning randomly.}
    \label{fig:attn_metric_heatmaps}
\end{figure*}

This approach ensures that only the projection channels contributing most significantly to the attention scores are retained. Compared to the L2 norm criterion from FFNs, this method better preserves the original computational outcomes. As demonstrated in~\fig{fig:attn_metric_heatmaps}, our proposed oracle criterion maintains an attention map much more similar to the original dense QK computation compared to those derived by L2 norm or random pruning.

\subsection{SVD Sparsity Estimator}
The oracle criteria demonstrate that adaptively sparsifying the backbone weights during fine-tuning can significantly reduce computation while maintaining model quality. However, computing these oracle patterns requires partial dense computation such as the gate and up projections for FFN intermediate activations and QK projections for QK norms, making it impractical and negating potential speedup benefits. To address this challenge, we introduce an efficient SVD-based low-rank sparsity estimator that dynamically selects channels with minimal overhead. Our method directly approximates the oracle criterion using top-k singular value decomposition (SVD) of the model weights, in contrast to prior approaches that employ learned low-rank predictors~\citep{dejavu, alizadeh2024flash, akhauri2024shadowllm}, which raise concerns about generalization across different datasets and tasks.

\vspace{-5pt}
\begin{algorithm}[H]
\caption{SVD Sparsity Estimator}
\label{algo:svd-algo}
\begin{minted}[
    framesep=10pt, 
    fontsize=\small, 
    python3=true,    % Enables Python-specific syntax highlighting
    escapeinside=||, % Allows inline LaTeX if needed
    breaklines=true  % Ensures long lines wrap properly
]{diff}
|\textcolor{teal}{# Assume input tensor x of shape (B, S, D1)}|
|\textcolor{teal}{# Weight W of shape (B, D1, D2), SVD rank of k}|

- # Compute activations with oracle
- out = torch.bmm(x, W)

+ # Compute low-rank SVD weights (saved offline)
+ U, S, V = torch.linalg.svd(W, full_matrices=False)
+ W_A = U[:, :k] @ torch.diag(S[:k]).sqrt()
+ W_B = torch.diag(S[:k]).sqrt() @ V[:k, :]

+ # Compute activations with loaded SVD estimator
+ out = torch.bmm(torch.bmm(x, W_A), W_B)

|\textcolor{teal}{# Obtain channel indices with corresponding metric}|
indices = metric(out)
\end{minted}
\end{algorithm}
\vspace{-10pt}

The core idea is straightforward: instead of training a predictor to map inputs to sparsity masks, we project inputs onto a low-rank SVD decomposition of the original weights and compute the oracle metric from these projected activations, as detailed in~\algo{algo:svd-algo}. This approach produces sparsity masks that closely match those obtained from the full model while maintaining efficiency. The SVD decomposed components are computed offline and loaded at the start of fine-tuning alongside the model weights. Notably, while low-rank module overheads can be significant in inference-time sparsity methods due to memory-bound execution~\citep{akhauri2024shadowllm, dejavu}, our approach introduces minimal overhead (less than 1\% of runtime) since the lightweight SVD projections are negligible compared to the matrix multiplications in the compute-bound LLM fine-tuning. This enables us to achieve dynamic sparsity with negligible computational overhead, preserving both fine-tuning efficiency and model performance.

\subsection{Sensitivity Analysis}

\myparagraph{Layer Sensitivity: Adaptive Sparsity Configuration}
The inherent contextual sparsity across layers in LLMs often varies significantly~\citep{dejavu, liu2024trainingfree}. Moreover, the importance of individual layers and their contributions to fine-tuning can differ substantially~\citep{gromov2024unreasonable, zheng2024learn}, necessitating layer-specific sparsity configurations for optimal performance. To determine these configurations, we conduct a systematic layer sensitivity analysis using a subset of the Commonsense Reasoning task proposed by~\citet{hu2023llmadapter} as a proxy. Our analysis evaluates how different sparsity ratios affect each layer's performance independently -- starting with a densely fine-tuned model, we progressively increase the sparsity ratio for each layer while keeping others dense, measuring performance each time to generate layer-specific sensitivity curves. 

\begin{figure}[t]
    \centering
    \includegraphics[width=.49\textwidth]{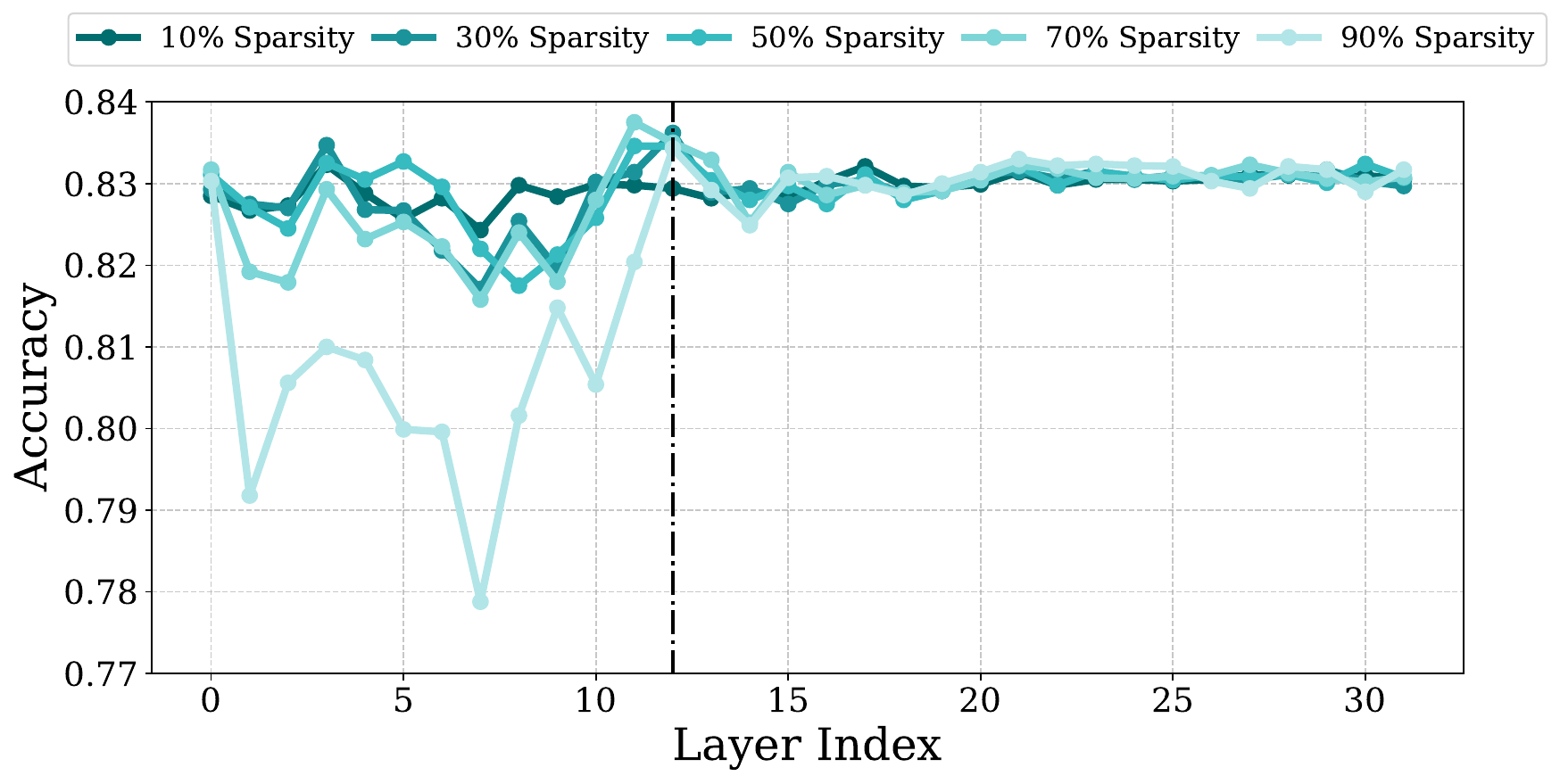}
    \caption{Sensitivity analysis on layer-wise sparsity of LLaMA2-7B.}
    \label{fig:layer-sensitivity}
\end{figure}

 The results for LLaMA2-7B, shown in~\fig{fig:layer-sensitivity}, reveal that deeper layers contain more redundant information and are more amenable to sparsification than shallower layers, aligning with inference-time observations from~\citet{gromov2024unreasonable}. These sensitivity metrics enable us to apply aggressive sparsity to deeper, resilient layers while maintaining shallower ones dense, optimizing the performance-efficiency trade-off during fine-tuning.

\myparagraph{Token Sensitivity: Context-Output Aware Sparsity}
The effectiveness of sparsity varies not only across layers but also across tokens within a sequence. In LLM fine-tuning, input sequences typically consist of a \textit{context} (the prefix tokens provided as input) and \textit{output} tokens (the target tokens used for loss computation). We find that applying uniform sparsity across all tokens degrades performance, as output tokens play a more critical role in optimization.

\begin{figure}[t]
    \centering
    \includegraphics[width=0.48\textwidth]{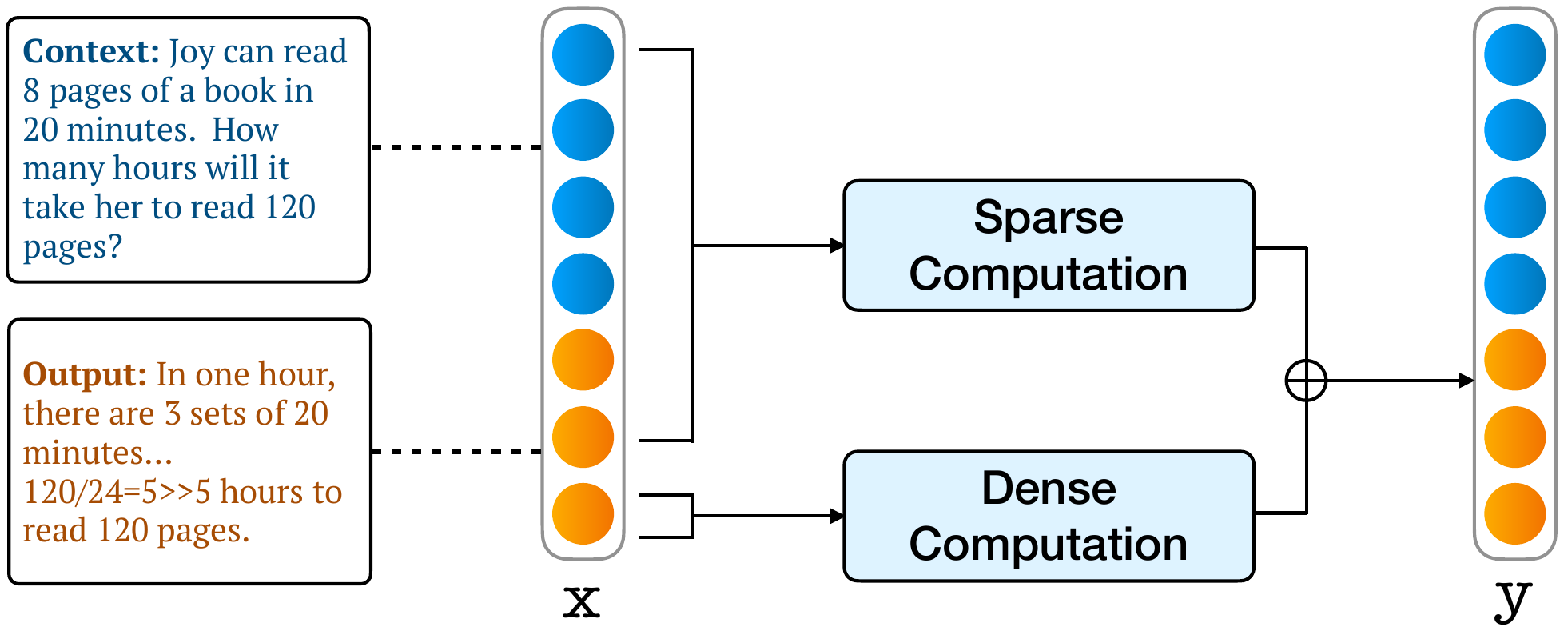}
    \caption{Output tokens go through the dense computation in our context-output aware sparsity strategy. The final outputs are gathered from both sparse and dense results.}
    \label{fig:context_ouptut}
\end{figure}

To address this, we propose a context-output aware sparsity strategy, selectively preserving dense computation for output tokens while applying sparsity to the context. This ensures that fine-tuning retains full expressiveness where it matters most while still benefiting from reduced computation. Unlike heuristic-based token importance sampling, our approach exploits a natural structural distinction -- context tokens are inherently less sensitive to precise weight updates than output tokens, as illustrated in~\fig{fig:context_ouptut}. This strategy significantly mitigates the reconstruction errors between sparse and dense training, particularly in early fine-tuning steps where maintaining gradient signal is crucial. This simple yet effective approach improves computational efficiency while preserving fine-tuning performance.


\myparagraph{Step Sensitivity: Progressive Sparse Fine-tuning}
To balance efficiency and model quality, we incorporate a hybrid approach in our fine-tuning process. Recent studies suggest that when doing incorporating sparsity in training, maintaining dense computations for a small portion of steps can significantly enhance final convergence~\citep{lu2023step, spdf, bambhaniya2024progressive} with minimal impact on overall speed-up. In our \method design, we allow the initial steps, up to a maximum of 10\% of the fine-tuning process, to remain dense. This approach ensures the model establishes a strong foundation early on while still benefiting from sparse training's efficiency in later stages. A detailed analysis of this hybrid approach's impact on performance and efficiency can be found in \ssect{ssect:ablation}.

\begin{table*}[t]
\renewcommand{\arraystretch}{1.1}
\small\centering
\begin{tabular}{lccccccccccc}
    \toprule
    & \#FLOPs & Speedup & Average & BoolQ & PIQA & SIQA & HellaS & WinoG & ARC-e & ARC-c & OBQA \\
    \midrule\midrule
    LLaMA2-7B & -- & -- & \textcolor{gray}{31.4} & \textcolor{gray}{51.0} & \textcolor{gray}{49.5} & \textcolor{gray}{32.4} & \textcolor{gray}{25.0} & \textcolor{gray}{20.4} & \textcolor{gray}{23.7} & \textcolor{gray}{22.1} & \textcolor{gray}{26.6} \\
    \midrule
    + LoRA & 100\% & 1.0$\times$ & 
    82.3 & 70.7 & 84.8 & 81.4 & 90.0 & 85.8 & 87.6 & 74.3 & 84.7 \\  
    + QLoRA & 100\% & 0.9$\times$ & 82.5 & 69.2 & 84.7 & 81.9 & 90.5 & 85.8 & 87.9 & 74.2 & 85.4\\
    + DoRA & 132\% & 0.7$\times$ & 81.7 & 71.4 & 84.7 & 81.1 & 90.0 & 85.2 & 87.3 & 72.8 & 84.2 \\
    + \textbf{\method} & 65\% & 1.3$\times$ & 81.8 & 69.7 & 84.3 & 80.8 & 88.4 & 86.0 & 86.7 & 73.4 & 84.0 \\
    
    \midrule\midrule
    
    LLaMA2-13B & -- & -- & \textcolor{gray}{35.0} & \textcolor{gray}{61.9} & \textcolor{gray}{49.8} & \textcolor{gray}{31.8} & \textcolor{gray}{25.6} & \textcolor{gray}{17.2} & \textcolor{gray}{33.3} & \textcolor{gray}{30.5} & \textcolor{gray}{29.6} \\
    \midrule
    + LoRA & 100\% & 1.0$\times$ & 84.7 & 72.0 & 86.7 & 82.2 & 91.0 & 89.0 & 90.9 & 80.5 & 85.6 \\
    + \textbf{\method} & 61\% & 1.3$\times$& 85.0 &  74.1 & 87.1 & 82.4 & 92.3 & 88.3 & 89.8 & 78.7 & 86.9 \\

    \midrule\midrule
    
    LLaMA3-8B & -- & -- & \textcolor{gray}{62.5} & \textcolor{gray}{66.1} & \textcolor{gray}{75.4} & \textcolor{gray}{53.8} & \textcolor{gray}{54.8} & \textcolor{gray}{42.1} & \textcolor{gray}{80.7} & \textcolor{gray}{67.3} & \textcolor{gray}{59.8} \\
    \midrule
    + LoRA & 100\% & 1.0$\times$ & 87.1 & 74.6 & 89.4 & 82.7 & 95.4 & 89.1 & 92.8 & 83.4 & 89.3 \\
    + QLoRA & 100\% & 0.9$\times$ & 87.1 & 74.3 & 89.3 & 83.1 & 95.3 & 88.7 & 92.9 & 83.8 & 89.4\\
    + DoRA & 132\% & 0.8$\times$ & 87.1 & 74.5 & 89.4 & 83.0 & 95.4 & 88.8 & 93.2 & 84.0 & 88.9 \\
    + \textbf{\method} & 65\% & 1.3$\times$  & 86.9 & 75.0 & 89.6 & 82.8 & 94.9 & 88.7 & 92.7 & 82.9 & 88.3 \\
    \bottomrule
\end{tabular}
\caption{\method delivers up to 1.3$\times$ speedup and reduces fine-tuning FLOPs by up to 39\%, while maintaining performance comparable to existing methods such as LoRA, QLoRA and DoRA on commonsense reasoning benchmarks.}
\label{tab:commonsense}
\end{table*}
\begin{table}[t]
\renewcommand{\arraystretch}{1.1}
\setlength{\tabcolsep}{2pt}
\small\centering
\begin{tabular}{lccccccc}
    \toprule
    & \#FL. & Spd. & Avg. & \makecell{GSM8K} & \makecell{SVAMP} & \makecell{MAWPS} \\
    \midrule\midrule
    LLaMA2-7B & -- & -- & \textcolor{gray}{2.6} & \textcolor{gray}{2.7} & \textcolor{gray}{3.1} & \textcolor{gray}{2.1} \\
    \midrule
    + LoRA & 100\% & 1.0$\times$ & 54.6 & 38.6 & 47.5 & 77.5 \\
    + QLoRA & 100\% & 0.9$\times$ & 55.0 & 36.2 & 49.7 & 79.1\\
    + DoRA & 132\% & 0.7$\times$  & 54.5 & 38.4 & 48.4 & 77.5\\
    + \textbf{\method} & 73\% & 1.2$\times$ & 53.7 & 37.6 & 46.4 & 77.9 \\
    \midrule\midrule
    LLaMA2-13B & -- & -- & \textcolor{gray}{13.4} & \textcolor{gray}{4.9} & \textcolor{gray}{18.8} & \textcolor{gray}{16.4} \\
    \midrule
    + LoRA & 100\% & 1.0$\times$ & 63.5 & 50.2 & 59.1 & 81.0 \\
    + \textbf{\method} & 70\% & 1.3$\times$ & 62.7 & 49.5 & 57.1 & 81.5 \\
    \midrule\midrule
    LLaMA3-8B & -- & -- & \textcolor{gray}{33.5} & \textcolor{gray}{25.0} & \textcolor{gray}{38.4} & \textcolor{gray}{37.0} \\
    \midrule
    + LoRA & 100\% & 1.0$\times$ & 81.0 & 71.8 & 80.3 & 90.9 \\
    + QLoRA & 100\% & 0.9$\times$ & 80.6 & 71.8 & 80.2 & 89.6\\
    + DoRA & 132\% & 0.8$\times$ & 81.0 & 72.5 & 79.3 & 91.0\\
    + \textbf{\method} & 46\% & 1.6$\times$ & 81.1 & 72.0 & 80.2 & 90.9 \\
    \bottomrule
\end{tabular}
\caption{\method offers up to 1.6$\times$ speedup and reduces fine-tuning FLOPs by up to 54\% on arithmetic reasoning tasks with accuracy comparable to existing methods.}
\label{tab:math}
\end{table}

\section{Experiments}
\label{sec:experiments}

\subsection{Setup}

\myparagraph{Benchmarks.}
We conduct experiments on five downstream tasks. The first set focuses on commonsense reasoning (referred to as CSR170K) and includes eight datasets: BoolQ~\citep{boolq}, PIQA~\citep{bisk2020piqa}, SIQA~\citep{siqa}, HellaSwag~\citep{hellaswag}, WinoGrande~\citep{sakaguchi2021winogrande}, ARC-Easy and ARC-Challenge~\citep{clark2018arc}, and OpenbookQA~\citep{obqa}. The second set focuses on arithmetic reasoning (referred to as Math10K) and includes three benchmarks: GSM8K~\citep{cobbe2021gsm}, MAWPS~\citep{mawps}, and SVAMP~\citep{svamp}\footnote{We exclude AQuA~\citep{aqua} since none of the methods in~\citet{hu2023llmadapter} achieve better-than-random performance (i.e., significantly above 20\% for a 5-choice multiple-choice task).}. Following the practices established by \citet{hu2023llmadapter} and \citet{dora}, we fine-tune our models on the combined training sets of all sub-tasks within each respective benchmark. We run each experiment five times, discard the highest and lowest performing runs, and report the average accuracy of the remaining three. We further assess the generality of our method on three additional tasks: sequence classification using the GLUE benchmark~\citep{wang2018glue}; instruction following, where we train on WizardLM~\citep{xu2024wizardlm} and evaluate on MT-Bench~\citep{zheng2023judging}; and code generation, where we train on CodeFeedback~\citep{chen2021evaluating} and test on HumanEval / HumanEval+~\citep{zheng2024opencodeinterpreter, evalplus}.

\myparagraph{Models.}
We use LLaMA2-7B/13B and LLaMA3-8B (Instruct) as our base models for fine-tuning. For instruction following and code generation benchmarks, we additionally use LLaMA3.1-8B. 

\myparagraph{Baselines.}
We compare our method with two PEFT methods, LoRA~\citep{lora} and DoRA~\citep{dora}. While SparseLoRA is built on top of LoRA, it can, in principle, be applied to any other PEFT method. We include model training details in Table~\ref{tab:dataset_details}. All PeFT methods only fine-tune the QKVO projections using a rank of 32, a scaling factor $\alpha$ of 64, and no dropout. Efficiency metrics are derived from an NVIDIA A6000 GPU.

\subsection{Main Results}

\tbl{tab:commonsense} demonstrates results on the CSR170K benchmarks, demonstrating that SparseLoRA significantly reduces computational requirements while maintaining accuracy comparable to LoRA. For instance, on LLaMA2-13B, SparseLoRA achieves an average accuracy of 85.0, outperforming LoRA's 84.7, while requiring only 61\% of its training cost with a 1.3$\times$ speedup in its training time. This trend extends to LLaMA3-8B, where SparseLoRA consistently reduces compute load while preserving competitive accuracy across tasks such as BoolQ, PIQA, SIQA, HellaSwag, WinoGrande, ARC, and OBQA.

\begin{table*}[h]
    \renewcommand{\arraystretch}{1.1}
    \setlength{\tabcolsep}{3.5pt}
    \small\centering
    \begin{tabular}{lccccccccccccc}
        \toprule
        & \#FLOPs & Speedup & Average & COLA & STS-B & MRPC & RTE & SST2 & QNLI & WNLI & MNLI & QQP\\
        \midrule
        LLaMA3-8B + LoRA & 100\% & 1.0$\times$ & 87.3 & 65.8 & 88.8 & 87.7 & 82.8 & 96.4 & 95.7 & 45.5 & 91.8 & 89.8 \\
        
        LLaMA3-8B + \textbf{\method} & 61\% & 1.3$\times$ &
        87.7 & 66.3 & 89.3 & 88.6 & 82.9 & 96.6 & 96.6 & 55.9 & 91.8 & 89.8\\
        

    
        \bottomrule
    \end{tabular}
    \caption{\method accelerates fine-tuning for sequence classification on the GLUE benchmark by 1.3 times.}
    \label{tab:glue}
\end{table*}

\begin{table*}[h]
    \renewcommand{\arraystretch}{1.1}
    \setlength{\tabcolsep}{4pt}
    \small\centering
    \begin{tabular}{lccccccccccc}
        \toprule
        & \#FLOPs & Speedup & Average & Coding & Extraction & Humanities & Math & Reasoning & Roleplay & STEM & Writing\\
        \midrule
        LLaMA3.1-8B & -- & -- &
        \textcolor{gray}{4.08} & \textcolor{gray}{2.15} & \textcolor{gray}{2.88} &
        \textcolor{gray}{4.50} & \textcolor{gray}{2.05} & \textcolor{gray}{2.60} & \textcolor{gray}{5.72} & \textcolor{gray}{7.90} & \textcolor{gray}{4.80} \\
        \midrule
        + LoRA & 100\% & 1.0$\times$ & 6.03 & 5.10 & 5.50 & 8.00 & 2.25 & 5.30 & 7.35 & 7.78 & 7.00 \\
        + \textbf{\method} & 53\% & 1.5$\times$ & 6.06 & 5.30 & 5.45 & 8.10 & 2.20 & 5.35 & 7.35 & 7.70 & 7.05 \\
        \bottomrule
    \end{tabular}
    \caption{\method delivers strong performance on instruction-following tasks in MT-Bench, achieving a 1.5$\times$ speedup while matching or exceeding LoRA across all categories.}
    \vspace{-4mm}
    \label{tab:mtbench}
\end{table*}

\tbl{tab:math} presents results on the Math10K benchmarks, further reinforcing SparseLoRA's compute–accuracy advantage. On LLaMA3-8B, SparseLoRA reduces training costs by 54\% and achieves a 1.6$\times$ speedup on LoRA while maintaining strong performance on SVAMP and MAWPS while slightly surpassing LoRA on GSM8K. These findings demonstrate that structured contextual sparsity can significantly reduce computational overhead in parameter-efficient fine-tuning without sacrificing performance in commonsense and mathematical reasoning tasks.

\myparagraph{Natural Language Understanding.}

We evaluate \method using LLaMA3-8B on sequence classification with a subset of GLUE benchmark~\citep{wang2018glue}. \tbl{tab:glue} shows that~\method maintains competitive performance to the dense baseline with a 1.3$\times$ speedup.

\myparagraph{Instruction Following.}

We evaluate \method using LLaMA3.1-8B on the task of instruction following by fine-tuning on a subset of the WizardLM dataset~\citep{xu2024wizardlm} and reporting scores across eight tasks in the MT-Bench dataset~\citep{zheng2023judging}. We use GPT-4 to assess the quality of model responses. ~\tbl{tab:mtbench} shows that~\method maintains competitive performance to the dense baseline while achieving up to a 1.5$\times$ speedup.

\myparagraph{Code Generation.}

We evaluate \method using LLaMA2-7B and LLaMA3.1-8B on code generation by fine-tuning on a subset of the CodeFeedback dataset~\citep{chen2021evaluating} and testing on the HumanEval benchmarks~\citep{zheng2024opencodeinterpreter, evalplus}. \tbl{tab:codefeedback} shows that~\method maintains competitive performance to the dense baseline while achieving up to a 1.3$\times$ speedup. 

\begin{table}[h]
    \renewcommand{\arraystretch}{1.1}
    \setlength{\tabcolsep}{2pt}
    \small\centering
    \begin{tabular}{lcccc}
        \toprule
        & \#FLOPs & Speedup & HumanEval & HumanEval+ \\
        \midrule\midrule
        LLaMA2-7B & -- & -- & \textcolor{gray}{3.6}  & \textcolor{gray}{3.0} \\
        \midrule 
        + LoRA & 100\% & 1.0$\times$ & 13.0 & 10.2 \\
        + \textbf{\method} & 73\% & 1.2$\times$ & 12.8 & 11.0 \\
        \midrule\midrule
        LLaMA3.1-8B & -- & -- & \textcolor{gray}{30.9}  & \textcolor{gray}{27.9} \\
        \midrule
        + LoRA & 100\% & 1.0$\times$ & 43.1 & 36.2 \\
        + \textbf{\method} & 66\% & 1.3$\times$ & 43.9 & 37.0 \\
        \bottomrule
    \end{tabular}
    \caption{\method accelerates fine-tuning for code generation by up to 1.3$\times$ while maintaining performance on HumanEval and HumanEval+ benchmarks.}
    \label{tab:codefeedback}
\end{table}


\myparagraph{Compatibility with PEFT Methods.}

\begin{table}[t]
\renewcommand{\arraystretch}{1.1}
\setlength{\tabcolsep}{4pt}
\small\centering
\begin{tabular}{lcccccc}
    \toprule
    & \multicolumn{3}{c}{CSR170K} & \multicolumn{3}{c}{Math10K} \\
    \cmidrule(lr){2-4} \cmidrule(lr){5-7}
     & \#FL. & Spd. & Acc. & \#FL. & Spd. & Acc. \\
    \midrule
    LLaMA3-8B & -- & -- & \textcolor{gray}{62.5}  & -- & -- & \textcolor{gray}{33.5} \\
    \midrule
    + QLoRA & 100\% & 1.0$\times$ & 87.1 & 100\% & $1.0\times$ & 80.6 \\
    + \textbf{SparseQLoRA} & 65\% & 1.2$\times$ & 86.9 & 60\%  & 1.3$\times$ & 80.8 \\
    \bottomrule
\end{tabular}
\caption{\method can be combined with existing PeFT approaches, such as QLoRA, to accelerate fine-tuning while maintaining memory savings.}
\label{tab:qlora-peft}
\end{table}

Techniques such as gradient checkpointing and quantization (\eg, QLoRA~\cite{dettmers2023qlora}, LoftQ~\citep{li2023loftq}) primarily aim to reduce memory usage but often increase runtime, as shown in Figure~\ref{fig:teaser}. These methods are therefore orthogonal and complementary to our approach. As in Table~\ref{tab:qlora-peft}, SparseLoRA can be combined with QLoRA to achieve both lower memory consumption and improved runtime efficiency.

\subsection{Analysis}
\label{ssect:ablation}


\myparagraph{SVD Sparsity Estimator.}

\begin{table}[h]
\renewcommand{\arraystretch}{1.1}
\setlength{\tabcolsep}{8pt}
\small\centering
\begin{tabular}{lcccc}
    \toprule
    & \#FLOPs & Runtime & Memory & Accuracy \\
    \midrule
    Oracle & -- & -- & -- & 81.4 \\
    SVD & 0.05\% & 0.8\% & 30MB & 81.1 \\
    \bottomrule
\end{tabular}
\caption{SVD sparsity estimator delivers near-oracle accuracy while introducing negligible computation and memory overheads, enabling \method to outperform LoRA in efficiency while matching its performance.}
\label{tab:svd_predictor}
\end{table}

The SVD sparsity estimator is key to \method’s ability to apply contextual sparsity with minimal impact on performance. As shown in \tbl{tab:svd_predictor}, our estimator achieves performance comparable to the Oracle method on the Math10K dataset, demonstrating its effectiveness. Using a rank 8 singular value decomposition of the base model weights, the SVD sparsity estimator is lightweight and training-free, adding only 0.8\% overhead to the end-to-end runtime.


\myparagraph{Effect of Output Token Splitting.}

We investigate the effectiveness of our context-output splitting strategy, which optimizes computational efficiency by preserving dense computation for output tokens while sparsifying the context tokens. To evaluate the impact of this design choice, we compare three configurations: (1) the baseline, which does not employ output token splitting, (2) our proposed approach with output token splitting, and (3) a control configuration, where a random subset of tokens is selected for dense computation, matching the number of output tokens in our approach. The results are presented in \tbl{tab:output_split}, which demonstrate that selectively preserving dense computation for output tokens consistently outperforms both the random selection and the baseline, highlighting the efficacy of our proposed method in improving computational efficiency without sacrificing performance.
\begin{table}[h]
\renewcommand{\arraystretch}{1.1}
\setlength{\tabcolsep}{5.5pt}
\small\centering
\begin{tabular}{lcccccc}
    \toprule
    & \multicolumn{3}{c}{CSR170K} & \multicolumn{3}{c}{Math10K} \\
    \cmidrule(lr){2-4} \cmidrule(lr){5-7}
    Sparsity & \#FL. & Spd. & Acc. & \#FL. & Spd. & Acc. \\
    \midrule
    All tokens & 100\% & 1.0$\times$ & 86.7 & 100\% & 1.0$\times$ & 47.1 \\
    Random & 102\% & 0.9$\times$ & 86.6 & 129\% & 0.8$\times$ & 70.9 \\
    Inputs only & 102\% & 0.9$\times$ & \textbf{86.9} & 129\% & 0.8$\times$ & \textbf{81.1} \\
    \bottomrule
\end{tabular}
\caption{Datasets with more output tokens (\eg, Math10K) are highly sensitive to sparsity on those tokens, leading to significant accuracy drops. Output-aware splitting preserves accuracy while still achieving strong runtime improvements.}
\label{tab:output_split}
\end{table}

\myparagraph{Uniform Sparsity without Sensitivity Analysis.}

We investigate the impact of applying sensitivity-guided layerwise sparsity compared to uniform sparsity across all model layers. Our experiments, conducted at various speedup targets, show that the sensitivity-aware approach—where sparsity ratios are adapted to each layer’s sensitivity—consistently outperforms the uniform sparsity baseline, as demonstrated in \tbl{tab:layerwise_sparsity}. This result underscores the importance of tailoring sparsification strategies to the unique sensitivity characteristics of each layer, rather than adopting a one-size-fits-all approach. Additionally, \fig{fig:layer-sensitivity} illustrates the varying sensitivity of sparsity across different layers, further validating the effectiveness of our layer-wise approach.
 \begin{table}[t]
\renewcommand{\arraystretch}{1.1}
    \setlength{\tabcolsep}{9pt}
    \small\centering
    \begin{tabular}{lccc}
        \toprule
        Sparsity Method& \#FLOPs & Speedup & Accuracy \\
        \midrule
        Uniform & 60\% & 1.1$\times$ & 80.3 \\
        Nonuniform & 60\% & 1.4$\times$ & \textbf{81.1}\\
        \midrule
        Uniform & 46\% & 1.5$\times$ & 80.2 \\
        Nonuniform & 46\% & 1.6$\times$ & \textbf{81.1}\\
        \midrule
        Uniform & 37\% & 1.6$\times$ & 79.5 \\
        Nonuniform & 37\% & 1.8$\times$ & \textbf{80.5}\\
        \bottomrule
        
    \end{tabular}
    \caption{On LLaMA3-8B with Math10K, we demonstrate the importance of layerwise sensitivity in sparsity allocation. At fixed FLOPs budgets, our non-uniform (sensitivity-aware) approach consistently achieves higher speedup and better accuracy compared to uniform sparsity. \method delivers lossless performance even at up to 1.6$\times$ speedup.}
    \label{tab:layerwise_sparsity}
\end{table}


\myparagraph{Pruning Criterion.}
We explore pruning criteria beyond L2 norm, such as methods based on weights \citep{sun2024wanda}. While these approaches show some promise, L2 norm remains the most effective method with simplicity. We compare L2 norm pruning with Wanda \citep{sun2024wanda} and random pruning, all using the oracle setting for FFNs. Additionally, we conduct an ablation study on attention projections, pruning heads and channels (same or different per head), as shown in~\tbl{tab:pruning_crtieria}. Our proposed attention norm performs the best.

\begin{table}[t]
\renewcommand{\arraystretch}{1.1}
\setlength{\tabcolsep}{3pt}
\small\centering
\begin{tabular}{lcccc}
    \toprule
    & \multicolumn{2}{c}{TopK Selection Per} & \multirow{2.5}{*}{\makecell{Metric}} & \multirow{2.5}{*}{\makecell{Accuracy \\ (Math10K)}} \\
    \cmidrule{2-3}
    & Channel & Head \\ 
    \midrule
    \multirow{4}{*}{\makecell{QK \\Criterion}} 
    & \checkmark & \xmark & Attention Norm & \textbf{80.7} \\ 
    & \xmark & \checkmark & Attention Norm & 79.6 \\ 
    & \checkmark & \xmark & L2 Norm & 79.8 \\ 
    & \checkmark & \xmark & Random & 79.1 \\ 
    \midrule
    \multirow{2}{*}{\makecell{VO \\Criterion}} 
    & -- & -- & L2Norm & \textbf{81.4} \\ 
    & -- & -- & Random & 79.6 \\ 
    \midrule
    \multirow{3}{*}{\makecell{FFN \\Criterion}} 
    & -- & -- & L2Norm & \textbf{81.4} \\ 
    & -- & -- & Wanda & 81.3 \\ 
    & -- & -- & Random & 78.6 \\ 
    \bottomrule
\end{tabular}
\caption{Comparison of pruning criteria for QK, VO, and FFN modules under 90\% uniform sparsity with token splitting at a 5\% step offset. All other components are computed densely. For each module, the selected metric: attention norm (channel-wise) for QK, and L2 norm for VO and FFN, achieves the highest accuracy, validating our design choices.}
\label{tab:pruning_crtieria}
\end{table}





\section{Conclusion}

We introduced \textit{\method} to accelerate fine-tuning through contextual sparsity using a lightweight, training-free \textit{SVD sparsity estimator}. By dynamically selecting sparse weights for loss and gradient computation, \method reduces computational cost by up to \textbf{2.2$\times$} and achieves up to a \textbf{1.6$\times$} speedup while maintaining accuracy across a wide range of benchmarks. 

\newpage
\myparagraph{Acknowledgment.}

We would like to thank Google-BAIR Commons, Google DeepMind, and POSCO HOLDINGS for their support of this research. We are also grateful to NVIDIA for providing GPU hardware. 

\section*{Impact Statement}

This paper presents an approach for compute-efficient fine-tuning of LLMs. There are many potential societal consequences of our work, none of which we feel must be specifically highlighted here.

\bibliography{reference}

\begin{thebibliography}{70}
\providecommand{\natexlab}[1]{#1}
\providecommand{\url}[1]{\texttt{#1}}
\expandafter\ifx\csname urlstyle\endcsname\relax
  \providecommand{\doi}[1]{doi: #1}\else
  \providecommand{\doi}{doi: \begingroup \urlstyle{rm}\Url}\fi

\bibitem[Akhauri et~al.(2024)Akhauri, AbouElhamayed, Dotzel, Zhang, Rush, Huda, and Abdelfattah]{akhauri2024shadowllm}
Akhauri, Y., AbouElhamayed, A.~F., Dotzel, J., Zhang, Z., Rush, A.~M., Huda, S., and Abdelfattah, M.
\newblock Shadowllm: Predictor-based contextual sparsity for large language models.
\newblock \emph{Conference on Empirical Methods in Natural Language Processing}, 2024.

\bibitem[Alizadeh et~al.(2024)Alizadeh, Mirzadeh, Belenko, Khatamifard, Cho, Del~Mundo, Rastegari, and Farajtabar]{alizadeh2024flash}
Alizadeh, K., Mirzadeh, S.~I., Belenko, D., Khatamifard, S., Cho, M., Del~Mundo, C.~C., Rastegari, M., and Farajtabar, M.
\newblock {LLM} in a flash: Efficient large language model inference with limited memory.
\newblock In \emph{Proceedings of the 62nd Annual Meeting of the Association for Computational Linguistics}, pp.\  12562--12584, 2024.

\bibitem[Bambhaniya et~al.(2024)Bambhaniya, Yazdanbakhsh, Subramanian, Kao, Agrawal, Evci, and Krishna]{bambhaniya2024progressive}
Bambhaniya, A.~R., Yazdanbakhsh, A., Subramanian, S., Kao, S.-C., Agrawal, S., Evci, U., and Krishna, T.
\newblock Progressive gradient flow for robust n:m sparsity training in transformers.
\newblock \emph{arXiv preprint arXiv: 2402.04744}, 2024.

\bibitem[Bisk et~al.(2020)Bisk, Zellers, Gao, Choi, et~al.]{bisk2020piqa}
Bisk, Y., Zellers, R., Gao, J., Choi, Y., et~al.
\newblock Piqa: Reasoning about physical commonsense in natural language.
\newblock In \emph{Proceedings of the AAAI conference on artificial intelligence}, volume~34, pp.\  7432--7439, 2020.

\bibitem[Chen et~al.(2025)Chen, He, Hu, Yuan, and Yuan]{chen2025celora}
Chen, G., He, Y., Hu, Y., Yuan, K., and Yuan, B.
\newblock Ce-lora: Computation-efficient lora fine-tuning for language models.
\newblock \emph{arXiv preprint arXiv: 2502.01378}, 2025.

\bibitem[Chen et~al.(2021)Chen, Tworek, Jun, Yuan, Pinto, Kaplan, Edwards, Burda, Joseph, Brockman, et~al.]{chen2021evaluating}
Chen, M., Tworek, J., Jun, H., Yuan, Q., Pinto, H. P. D.~O., Kaplan, J., Edwards, H., Burda, Y., Joseph, N., Brockman, G., et~al.
\newblock Evaluating large language models trained on code.
\newblock \emph{arXiv preprint arXiv:2107.03374}, 2021.

\bibitem[Chen et~al.(2024)Chen, Qian, Tang, Lai, Liu, Han, and Jia]{longlora}
Chen, Y., Qian, S., Tang, H., Lai, X., Liu, Z., Han, S., and Jia, J.
\newblock Longlora: Efficient fine-tuning of long-context large language models.
\newblock In \emph{The International Conference on Learning Representations (ICLR)}, 2024.

\bibitem[Clark et~al.(2019)Clark, Lee, Chang, Kwiatkowski, Collins, and Toutanova]{boolq}
Clark, C., Lee, K., Chang, M.-W., Kwiatkowski, T., Collins, M., and Toutanova, K.
\newblock {B}ool{Q}: Exploring the surprising difficulty of natural yes/no questions.
\newblock In \emph{Proceedings of the 2019 Conference of the North {A}merican Chapter of the Association for Computational Linguistics: Human Language Technologies}, pp.\  2924--2936, 2019.

\bibitem[Clark et~al.(2018)Clark, Cowhey, Etzioni, Khot, Sabharwal, Schoenick, and Tafjord]{clark2018arc}
Clark, P., Cowhey, I., Etzioni, O., Khot, T., Sabharwal, A., Schoenick, C., and Tafjord, O.
\newblock Think you have solved question answering? try arc, the ai2 reasoning challenge.
\newblock \emph{arXiv preprint arXiv: 1803.05457}, 2018.

\bibitem[Cobbe et~al.(2021)Cobbe, Kosaraju, Bavarian, Chen, Jun, Kaiser, Plappert, Tworek, Hilton, Nakano, Hesse, and Schulman]{cobbe2021gsm}
Cobbe, K., Kosaraju, V., Bavarian, M., Chen, M., Jun, H., Kaiser, L., Plappert, M., Tworek, J., Hilton, J., Nakano, R., Hesse, C., and Schulman, J.
\newblock Training verifiers to solve math word problems.
\newblock \emph{arXiv preprint arXiv: 2110.14168}, 2021.

\bibitem[Dettmers et~al.(2023)Dettmers, Pagnoni, Holtzman, and Zettlemoyer]{dettmers2023qlora}
Dettmers, T., Pagnoni, A., Holtzman, A., and Zettlemoyer, L.
\newblock {QL}o{RA}: Efficient finetuning of quantized {LLM}s.
\newblock In \emph{Thirty-seventh Conference on Neural Information Processing Systems}, 2023.

\bibitem[Frantar \& Alistarh(2023)Frantar and Alistarh]{frantar2023sparsegpt}
Frantar, E. and Alistarh, D.
\newblock Sparsegpt: Massive language models can be accurately pruned in one-shot.
\newblock In \emph{International Conference on Machine Learning}, volume 202, pp.\  10323--10337. PMLR, 2023.

\bibitem[Gromov et~al.(2024)Gromov, Tirumala, Shapourian, Glorioso, and Roberts]{gromov2024unreasonable}
Gromov, A., Tirumala, K., Shapourian, H., Glorioso, P., and Roberts, D.~A.
\newblock The unreasonable ineffectiveness of the deeper layers.
\newblock \emph{arXiv preprint arXiv: 2403.17887}, 2024.

\bibitem[Han et~al.(2015)Han, Pool, Tran, and Dally]{han2015learning}
Han, S., Pool, J., Tran, J., and Dally, W.
\newblock Learning both weights and connections for efficient neural network.
\newblock In \emph{Advances in Neural Information Processing Systems (NIPS)}, pp.\  1135--1143, 2015.

\bibitem[Han et~al.(2016{\natexlab{a}})Han, Liu, Mao, Pu, Pedram, Horowitz, and Dally]{han2016eie}
Han, S., Liu, X., Mao, H., Pu, J., Pedram, A., Horowitz, M.~A., and Dally, W.~J.
\newblock Eie: Efficient inference engine on compressed deep neural network.
\newblock \emph{International Conference on Computer Architecture (ISCA)}, 2016{\natexlab{a}}.

\bibitem[Han et~al.(2016{\natexlab{b}})Han, Mao, and Dally]{han2015deep_compression}
Han, S., Mao, H., and Dally, W.~J.
\newblock Deep compression: Compressing deep neural networks with pruning, trained quantization and huffman coding.
\newblock \emph{International Conference on Learning Representations (ICLR)}, 2016{\natexlab{b}}.

\bibitem[Hayou et~al.(2024)Hayou, Ghosh, and Yu]{hayou2024loraplus}
Hayou, S., Ghosh, N., and Yu, B.
\newblock Lora+: Efficient low rank adaptation of large models.
\newblock In \emph{International Conference on Machine Learning}, 2024.

\bibitem[Hu et~al.(2022)Hu, Shen, Wallis, Allen{-}Zhu, Li, Wang, Wang, and Chen]{lora}
Hu, E.~J., Shen, Y., Wallis, P., Allen{-}Zhu, Z., Li, Y., Wang, S., Wang, L., and Chen, W.
\newblock Lora: Low-rank adaptation of large language models.
\newblock In \emph{The Tenth International Conference on Learning Representations}, 2022.

\bibitem[Hu et~al.(2023)Hu, Wang, Lan, Xu, Lim, Bing, Xu, Poria, and Lee]{hu2023llmadapter}
Hu, Z., Wang, L., Lan, Y., Xu, W., Lim, E., Bing, L., Xu, X., Poria, S., and Lee, R.~K.
\newblock Llm-adapters: An adapter family for parameter-efficient fine-tuning of large language models.
\newblock In \emph{Proceedings of the 2023 Conference on Empirical Methods in Natural Language Processing}, pp.\  5254--5276, 2023.

\bibitem[Jaiswal et~al.(2024)Jaiswal, Yin, Zhang, Liu, Zhao, Tian, and Wang]{jaiswal2024welore}
Jaiswal, A., Yin, L., Zhang, Z., Liu, S., Zhao, J., Tian, Y., and Wang, Z.
\newblock From galore to welore: How low-rank weights non-uniformly emerge from low-rank gradients.
\newblock \emph{arXiv preprint arXiv: 2407.11239}, 2024.

\bibitem[Koncel-Kedziorski et~al.(2016)Koncel-Kedziorski, Roy, Amini, Kushman, and Hajishirzi]{mawps}
Koncel-Kedziorski, R., Roy, S., Amini, A., Kushman, N., and Hajishirzi, H.
\newblock {MAWPS}: A math word problem repository.
\newblock In \emph{Proceedings of the 2016 Conference of the North {A}merican Chapter of the Association for Computational Linguistics: Human Language Technologies}, pp.\  1152--1157, 2016.

\bibitem[Kopiczko et~al.(2024)Kopiczko, Blankevoort, and Asano]{kopiczko2024vera}
Kopiczko, D.~J., Blankevoort, T., and Asano, Y.~M.
\newblock Vera: Vector-based random matrix adaptation.
\newblock In \emph{International Conference on Learning Representations}, 2024.

\bibitem[Lee et~al.(2024)Lee, Lee, Zhang, Tiwari, and Mirhoseini]{lee2024cats}
Lee, J.-Y., Lee, D., Zhang, G., Tiwari, M., and Mirhoseini, A.
\newblock Cats: Contextually-aware thresholding for sparsity in large language models.
\newblock \emph{arXiv preprint arXiv: 2404.08763}, 2024.

\bibitem[Li et~al.(2022)Li, Luo, Tan, Wang, Huang, Li, and Bai]{liijcai2022p586}
Li, Y., Luo, F., Tan, C., Wang, M., Huang, S., Li, S., and Bai, J.
\newblock Parameter-efficient sparsity for large language models fine-tuning.
\newblock In \emph{Proceedings of the Thirty-First International Joint Conference on Artificial Intelligence, {IJCAI-22}}, 2022.

\bibitem[Li et~al.(2023{\natexlab{a}})Li, Yu, Liang, He, Karampatziakis, Chen, and Zhao]{li2023loftq}
Li, Y., Yu, Y., Liang, C., He, P., Karampatziakis, N., Chen, W., and Zhao, T.
\newblock Loftq: Lora-fine-tuning-aware quantization for large language models.
\newblock \emph{International Conference on Learning Representations}, 2023{\natexlab{a}}.

\bibitem[Li et~al.(2023{\natexlab{b}})Li, You, Bhojanapalli, Li, Rawat, Reddi, Ye, Chern, Yu, Guo, and Kumar]{li2023lazyneurons}
Li, Z., You, C., Bhojanapalli, S., Li, D., Rawat, A.~S., Reddi, S.~J., Ye, K., Chern, F., Yu, F.~X., Guo, R., and Kumar, S.
\newblock The lazy neuron phenomenon: On emergence of activation sparsity in transformers.
\newblock In \emph{International Conference on Learning Representations}, 2023{\natexlab{b}}.

\bibitem[Ling et~al.(2017)Ling, Yogatama, Dyer, and Blunsom]{aqua}
Ling, W., Yogatama, D., Dyer, C., and Blunsom, P.
\newblock Program induction by rationale generation: Learning to solve and explain algebraic word problems.
\newblock In \emph{Proceedings of the 55th Annual Meeting of the Association for Computational Linguistics}, 2017.

\bibitem[Liu et~al.(2023{\natexlab{a}})Liu, Xia, Wang, and Zhang]{evalplus}
Liu, J., Xia, C.~S., Wang, Y., and Zhang, L.
\newblock Is your code generated by chat{GPT} really correct? rigorous evaluation of large language models for code generation.
\newblock In \emph{Thirty-seventh Conference on Neural Information Processing Systems}, 2023{\natexlab{a}}.
\newblock URL \url{https://openreview.net/forum?id=1qvx610Cu7}.

\bibitem[Liu et~al.(2024{\natexlab{a}})Liu, Ponnusamy, Cai, Guo, Kim, and Athiwaratkun]{liu2024trainingfree}
Liu, J., Ponnusamy, P., Cai, T., Guo, H., Kim, Y., and Athiwaratkun, B.
\newblock Training-free activation sparsity in large language models.
\newblock \emph{arXiv preprint arXiv: 2408.14690}, 2024{\natexlab{a}}.

\bibitem[Liu et~al.(2024{\natexlab{b}})Liu, Wang, Yin, Molchanov, Wang, Cheng, and Chen]{dora}
Liu, S.-Y., Wang, C.-Y., Yin, H., Molchanov, P., Wang, Y.-C.~F., Cheng, K.-T., and Chen, M.-H.
\newblock {D}o{RA}: Weight-decomposed low-rank adaptation.
\newblock In \emph{Proceedings of the 41st International Conference on Machine Learning}, volume 235, pp.\  32100--32121. PMLR, 2024{\natexlab{b}}.

\bibitem[Liu et~al.(2024{\natexlab{c}})Liu, Qiu, Feng, Xiu, Xue, Yu, Feng, Liu, Heo, Peng, Wen, Black, Weller, and Sch{\"{o}}lkopf]{liu2024butterflyoft}
Liu, W., Qiu, Z., Feng, Y., Xiu, Y., Xue, Y., Yu, L., Feng, H., Liu, Z., Heo, J., Peng, S., Wen, Y., Black, M.~J., Weller, A., and Sch{\"{o}}lkopf, B.
\newblock Parameter-efficient orthogonal finetuning via butterfly factorization.
\newblock In \emph{The Twelfth International Conference on Learning Representations}, 2024{\natexlab{c}}.

\bibitem[Liu et~al.(2023{\natexlab{b}})Liu, Wang, Dao, Zhou, Yuan, Song, Shrivastava, Zhang, Tian, R{\'{e}}, and Chen]{dejavu}
Liu, Z., Wang, J., Dao, T., Zhou, T., Yuan, B., Song, Z., Shrivastava, A., Zhang, C., Tian, Y., R{\'{e}}, C., and Chen, B.
\newblock Deja vu: Contextual sparsity for efficient llms at inference time.
\newblock In \emph{International Conference on Machine Learning}. PMLR, 2023{\natexlab{b}}.

\bibitem[Lu et~al.(2023)Lu, Agrawal, Subramanian, Rybakov, De~Sa, and Yazdanbakhsh]{lu2023step}
Lu, Y., Agrawal, S., Subramanian, S., Rybakov, O., De~Sa, C., and Yazdanbakhsh, A.
\newblock {STEP}: Learning {N}:{M} structured sparsity masks from scratch with precondition.
\newblock In \emph{Proceedings of the 40th International Conference on Machine Learning}, volume 202, pp.\  22812--22824. PMLR, 2023.

\bibitem[Ma et~al.(2024)Ma, Chen, Wang, Xu, Li, Sun, Zhu, Fan, and Yu]{ma2024sparsityaccelerated}
Ma, D., Chen, L., Wang, P., Xu, H., Li, H., Sun, L., Zhu, S., Fan, S., and Yu, K.
\newblock Sparsity-accelerated training for large language models.
\newblock \emph{Annual Meeting of the Association for Computational Linguistics}, 2024.

\bibitem[Meng et~al.(2024)Meng, Wang, and Zhang]{meng2024pissa}
Meng, F., Wang, Z., and Zhang, M.
\newblock Pissa: Principal singular values and singular vectors adaptation of large language models.
\newblock \emph{arXiv preprint arXiv: 2404.02948}, 2024.

\bibitem[Mihaylov et~al.(2018)Mihaylov, Clark, Khot, and Sabharwal]{obqa}
Mihaylov, T., Clark, P., Khot, T., and Sabharwal, A.
\newblock Can a suit of armor conduct electricity? a new dataset for open book question answering.
\newblock In \emph{Proceedings of the 2018 Conference on Empirical Methods in Natural Language Processing}, pp.\  2381--2391, 2018.

\bibitem[Mirzadeh et~al.(2023)Mirzadeh, Alizadeh-Vahid, Mehta, Mundo, Tuzel, Samei, Rastegari, and Farajtabar]{mirzadeh2023relu}
Mirzadeh, I., Alizadeh-Vahid, K., Mehta, S., Mundo, C. C.~D., Tuzel, O., Samei, G., Rastegari, M., and Farajtabar, M.
\newblock Relu strikes back: Exploiting activation sparsity in large language models.
\newblock \emph{International Conference on Learning Representations}, 2023.

\bibitem[Mozaffari et~al.(2024)Mozaffari, Yazdanbakhsh, Zhang, and Dehnavi]{mozaffari2024slope}
Mozaffari, M., Yazdanbakhsh, A., Zhang, Z., and Dehnavi, M.~M.
\newblock Slope: Double-pruned sparse plus lazy low-rank adapter pretraining of llms.
\newblock \emph{arXiv preprint arXiv: 2405.16325}, 2024.

\bibitem[Pan et~al.(2024)Pan, Liu, Diao, Pi, Zhang, Han, and Zhang]{pan2024lisa}
Pan, R., Liu, X., Diao, S., Pi, R., Zhang, J., Han, C., and Zhang, T.
\newblock Lisa: layerwise importance sampling for memory-efficient large language model fine-tuning.
\newblock \emph{Advances in Neural Information Processing Systems}, 2024.

\bibitem[Patel et~al.(2021)Patel, Bhattamishra, and Goyal]{svamp}
Patel, A., Bhattamishra, S., and Goyal, N.
\newblock Are {NLP} models really able to solve simple math word problems?
\newblock In \emph{Proceedings of the 2021 Conference of the North American Chapter of the Association for Computational Linguistics: Human Language Technologies}, pp.\  2080--2094, 2021.

\bibitem[Qiu et~al.(2023)Qiu, Liu, Feng, Xue, Feng, Liu, Zhang, Weller, and Sch{\"o}lkopf]{qiu2023oft}
Qiu, Z., Liu, W., Feng, H., Xue, Y., Feng, Y., Liu, Z., Zhang, D., Weller, A., and Sch{\"o}lkopf, B.
\newblock Controlling text-to-image diffusion by orthogonal finetuning.
\newblock \emph{Advances in Neural Information Processing Systems}, 36:\penalty0 79320--79362, 2023.

\bibitem[Saab et~al.(2024)Saab, Tu, Weng, Tanno, Stutz, Wulczyn, Zhang, Strother, Park, Vedadi, et~al.]{saab2024medgemini}
Saab, K., Tu, T., Weng, W.-H., Tanno, R., Stutz, D., Wulczyn, E., Zhang, F., Strother, T., Park, C., Vedadi, E., et~al.
\newblock {Capabilities of Gemini Models in Medicine}.
\newblock \emph{arXiv:2404.18416}, 2024.

\bibitem[Sakaguchi et~al.(2021)Sakaguchi, Bras, Bhagavatula, and Choi]{sakaguchi2021winogrande}
Sakaguchi, K., Bras, R.~L., Bhagavatula, C., and Choi, Y.
\newblock Winogrande: An adversarial winograd schema challenge at scale.
\newblock \emph{Communications of the ACM}, 64\penalty0 (9):\penalty0 99--106, 2021.

\bibitem[Sap et~al.(2019)Sap, Rashkin, Chen, Le~Bras, and Choi]{siqa}
Sap, M., Rashkin, H., Chen, D., Le~Bras, R., and Choi, Y.
\newblock Social {IQ}a: Commonsense reasoning about social interactions.
\newblock In \emph{Proceedings of the 2019 Conference on Empirical Methods in Natural Language Processing and the 9th International Joint Conference on Natural Language Processing (EMNLP-IJCNLP)}, pp.\  4463--4473, 2019.

\bibitem[Shi et~al.(2023)Shi, Gai, Darrell, and Wang]{shi2023toast}
Shi, B., Gai, S., Darrell, T., and Wang, X.
\newblock Toast: Transfer learning via attention steering.
\newblock \emph{arXiv preprint arXiv: 2305.15542}, 2023.

\bibitem[Song et~al.(2024{\natexlab{a}})Song, Han, Zhang, Hu, Shi, Li, Chen, Liu, Li, Yang, and Sun]{song2024prosparse}
Song, C., Han, X., Zhang, Z., Hu, S., Shi, X., Li, K., Chen, C., Liu, Z., Li, G., Yang, T., and Sun, M.
\newblock Prosparse: Introducing and enhancing intrinsic activation sparsity within large language models.
\newblock \emph{arXiv preprint arXiv: 2402.13516}, 2024{\natexlab{a}}.

\bibitem[Song et~al.(2023)Song, Mi, Xie, and Chen]{song2023powerinfer}
Song, Y., Mi, Z., Xie, H., and Chen, H.
\newblock Powerinfer: Fast large language model serving with a consumer-grade gpu, 2023.

\bibitem[Song et~al.(2024{\natexlab{b}})Song, Xie, Zhang, Wen, Ma, Mi, and Chen]{song2024turbo}
Song, Y., Xie, H., Zhang, Z., Wen, B., Ma, L., Mi, Z., and Chen, H.
\newblock Turbo sparse: Achieving llm sota performance with minimal activated parameters.
\newblock \emph{arXiv preprint arXiv: 2406.05955}, 2024{\natexlab{b}}.

\bibitem[Sun et~al.(2024)Sun, Liu, Bair, and Kolter]{sun2024wanda}
Sun, M., Liu, Z., Bair, A., and Kolter, J.~Z.
\newblock A simple and effective pruning approach for large language models.
\newblock In \emph{The Twelfth International Conference on Learning Representations}, 2024.

\bibitem[Team(2024{\natexlab{a}})]{team2024gemma}
Team, G.
\newblock Gemma: Open models based on gemini research and technology.
\newblock \emph{arXiv preprint arXiv: 2403.08295}, 2024{\natexlab{a}}.

\bibitem[Team(2024{\natexlab{b}})]{team2024gemma2}
Team, G.
\newblock Gemma 2: Improving open language models at a practical size.
\newblock \emph{arXiv preprint arXiv: 2408.00118}, 2024{\natexlab{b}}.

\bibitem[Team(2023)]{touvron2023llama2}
Team, L.
\newblock Llama 2: Open foundation and fine-tuned chat models.
\newblock \emph{arXiv preprint arXiv: 2307.09288}, 2023.

\bibitem[Team(2024{\natexlab{c}})]{dubey2024llama3}
Team, L.
\newblock The llama 3 herd of models.
\newblock \emph{arXiv preprint arXiv: 2407.21783}, 2024{\natexlab{c}}.

\bibitem[Thangarasa et~al.(2023)Thangarasa, Gupta, Marshall, Li, Leong, DeCoste, Lie, and Saxena]{spdf}
Thangarasa, V., Gupta, A., Marshall, W., Li, T., Leong, K., DeCoste, D., Lie, S., and Saxena, S.
\newblock {SPDF:} sparse pre-training and dense fine-tuning for large language models.
\newblock In \emph{Uncertainty in Artificial Intelligence}, volume 216, pp.\  2134--2146. PMLR, 2023.

\bibitem[Wang et~al.(2018)Wang, Singh, Michael, Hill, Levy, and Bowman]{wang2018glue}
Wang, A., Singh, A., Michael, J., Hill, F., Levy, O., and Bowman, S.~R.
\newblock Glue: A multi-task benchmark and analysis platform for natural language understanding.
\newblock \emph{arXiv preprint arXiv:1804.07461}, 2018.

\bibitem[Wang et~al.(2024)Wang, Yu, and Li]{wang2024loraga}
Wang, S., Yu, L., and Li, J.
\newblock Lora-ga: Low-rank adaptation with gradient approximation.
\newblock \emph{arXiv preprint arXiv: 2407.05000}, 2024.

\bibitem[Wang \& Liang(2024)Wang and Liang]{wang2024lorapro}
Wang, Z. and Liang, J.
\newblock Lora-pro: Are low-rank adapters properly optimized?
\newblock \emph{arXiv preprint arXiv: 2407.18242}, 2024.

\bibitem[Xiao et~al.(2023)Xiao, Lin, Seznec, Wu, Demouth, and Han]{xiao2023smoothquant}
Xiao, G., Lin, J., Seznec, M., Wu, H., Demouth, J., and Han, S.
\newblock {S}mooth{Q}uant: Accurate and efficient post-training quantization for large language models.
\newblock In \emph{Proceedings of the 40th International Conference on Machine Learning}, 2023.

\bibitem[Xu et~al.(2024)Xu, Sun, Zheng, Geng, Zhao, Feng, Tao, Lin, and Jiang]{xu2024wizardlm}
Xu, C., Sun, Q., Zheng, K., Geng, X., Zhao, P., Feng, J., Tao, C., Lin, Q., and Jiang, D.
\newblock Wizardlm: Empowering large pre-trained language models to follow complex instructions.
\newblock In \emph{The Twelfth International Conference on Learning Representations}, 2024.

\bibitem[Xue et~al.(2024)Xue, Song, Mi, Chen, Xia, and Chen]{xue2024powerinfer2}
Xue, Z., Song, Y., Mi, Z., Chen, L., Xia, Y., and Chen, H.
\newblock Powerinfer-2: Fast large language model inference on a smartphone.
\newblock \emph{arXiv preprint arXiv: 2406.06282}, 2024.

\bibitem[Yang et~al.(2024)Yang, Leng, Guo, Zhao, Nakada, Zhang, Yao, and Chen]{yang2024sft}
Yang, X., Leng, J., Guo, G., Zhao, J., Nakada, R., Zhang, L., Yao, H., and Chen, B.
\newblock S\${\textasciicircum}\{2\}\${FT}: Efficient, scalable and generalizable {LLM} fine-tuning by structured sparsity.
\newblock In \emph{The Thirty-eighth Annual Conference on Neural Information Processing Systems}, 2024.

\bibitem[Zellers et~al.(2019)Zellers, Holtzman, Bisk, Farhadi, and Choi]{hellaswag}
Zellers, R., Holtzman, A., Bisk, Y., Farhadi, A., and Choi, Y.
\newblock Hellaswag: Can a machine really finish your sentence?
\newblock In \emph{Proceedings of the 57th Conference of the Association for Computational Linguistics}, pp.\  4791--4800, 2019.

\bibitem[Zhang et~al.(2024{\natexlab{a}})Zhang, Han, Liu, Zhou, Lu, Qiao, Li, and Gao]{zhang2024llama}
Zhang, R., Han, J., Liu, C., Zhou, A., Lu, P., Qiao, Y., Li, H., and Gao, P.
\newblock {LLaMA-Adapter: Efficient Fine-tuning of Language Models with Zero-init Attention}.
\newblock In \emph{International Conference on Learning Representations (ICLR)}, 2024{\natexlab{a}}.

\bibitem[Zhang et~al.(2024{\natexlab{b}})Zhang, Jaiswal, Yin, Liu, Zhao, Tian, and Wang]{zhang2024qgalore}
Zhang, Z., Jaiswal, A., Yin, L., Liu, S., Zhao, J., Tian, Y., and Wang, Z.
\newblock Q-galore: Quantized galore with int4 projection and layer-adaptive low-rank gradients.
\newblock \emph{arXiv preprint arXiv: 2407.08296}, 2024{\natexlab{b}}.

\bibitem[Zhang et~al.(2024{\natexlab{c}})Zhang, Song, Yu, Han, Lin, Xiao, Song, Liu, Mi, and Sun]{zhang2024relu2}
Zhang, Z., Song, Y., Yu, G., Han, X., Lin, Y., Xiao, C., Song, C., Liu, Z., Mi, Z., and Sun, M.
\newblock Relu$^2$ wins: Discovering efficient activation functions for sparse llms.
\newblock \emph{arXiv preprint arXiv: 2402.03804}, 2024{\natexlab{c}}.

\bibitem[Zhao et~al.(2024)Zhao, Zhang, Chen, Wang, Anandkumar, and Tian]{zhao2024galore}
Zhao, J., Zhang, Z., Chen, B., Wang, Z., Anandkumar, A., and Tian, Y.
\newblock {G}a{L}ore: Memory-efficient {LLM} training by gradient low-rank projection.
\newblock In \emph{Proceedings of the 41st International Conference on Machine Learning}, volume 235, pp.\  61121--61143. PMLR, 2024.

\bibitem[Zheng et~al.(2024{\natexlab{a}})Zheng, Bai, Liu, Mao, Chen, Lai, and Prakash]{zheng2024learn}
Zheng, H., Bai, X., Liu, X., Mao, Z., Chen, B., Lai, F., and Prakash, A.
\newblock Learn to be efficient: Build structured sparsity in large language models.
\newblock In \emph{The Thirty-eighth Annual Conference on Neural Information Processing Systems}, 2024{\natexlab{a}}.

\bibitem[Zheng et~al.(2023)Zheng, Chiang, Sheng, Zhuang, Wu, Zhuang, Lin, Li, Li, Xing, et~al.]{zheng2023judging}
Zheng, L., Chiang, W.-L., Sheng, Y., Zhuang, S., Wu, Z., Zhuang, Y., Lin, Z., Li, Z., Li, D., Xing, E., et~al.
\newblock Judging llm-as-a-judge with mt-bench and chatbot arena.
\newblock \emph{Advances in Neural Information Processing Systems}, 36:\penalty0 46595--46623, 2023.

\bibitem[Zheng et~al.(2024{\natexlab{b}})Zheng, Zhang, Shen, Liu, Lin, Fu, Chen, and Yue]{zheng2024opencodeinterpreter}
Zheng, T., Zhang, G., Shen, T., Liu, X., Lin, B.~Y., Fu, J., Chen, W., and Yue, X.
\newblock Opencodeinterpreter: Integrating code generation with execution and refinement.
\newblock \emph{arXiv preprint arXiv:2402.14658}, 2024{\natexlab{b}}.

\bibitem[Zhou et~al.(2021)Zhou, Zhang, Chen, Diao, and Zhang]{zhou2021efficient}
Zhou, X., Zhang, W., Chen, Z., Diao, S., and Zhang, T.
\newblock Efficient neural network training via forward and backward propagation sparsification.
\newblock \emph{Advances in Neural Information Processing Systems}, 34, 2021.

\end{thebibliography}
\bibliographystyle{icml2025}

\newpage
\appendix
\onecolumn
\section{Appendix}
\subsection{Analysis of Pruning at Attention Head Level in Inference and Fine-tuning}
\label{append:attn_head}
\begin{figure*}[h]
    \centering
    \begin{subfigure}{0.4\textwidth}
        \centering
        \includegraphics[width=\textwidth]{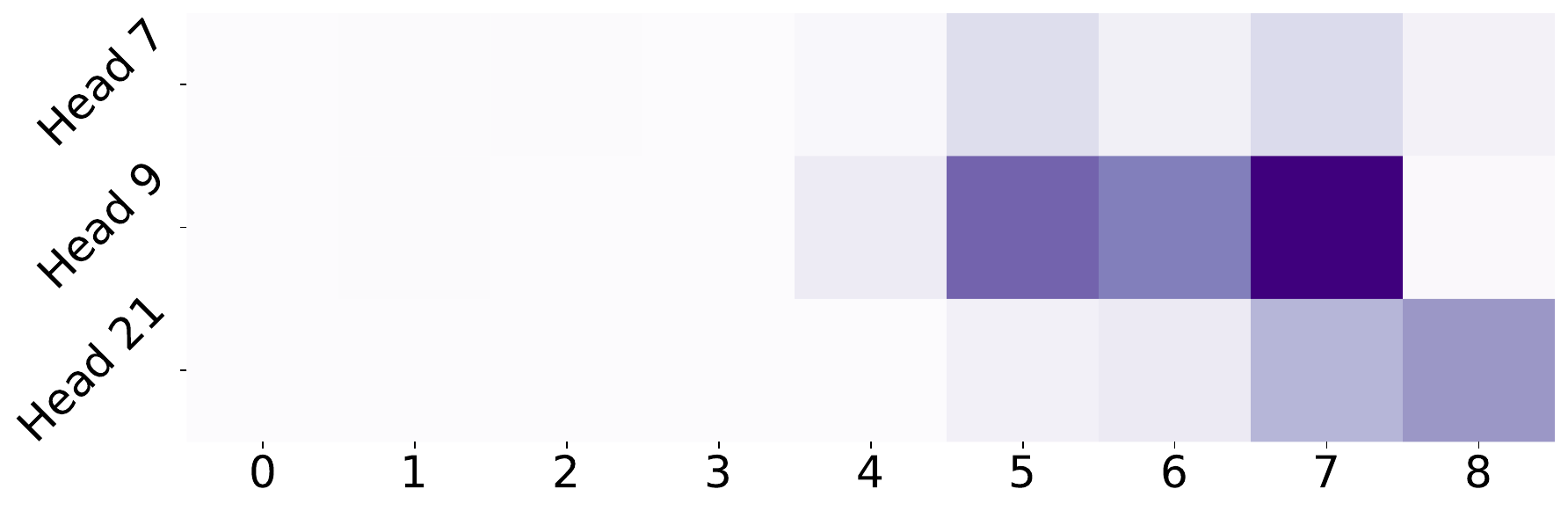}
        \caption{last row in \fig{fig:attn-head-multi}'s attention scores visualization}
        \label{fig:attn-head-single}
    \end{subfigure}\hfill
    \begin{subfigure}{0.55\textwidth}
        \centering
        \includegraphics[width=\textwidth]{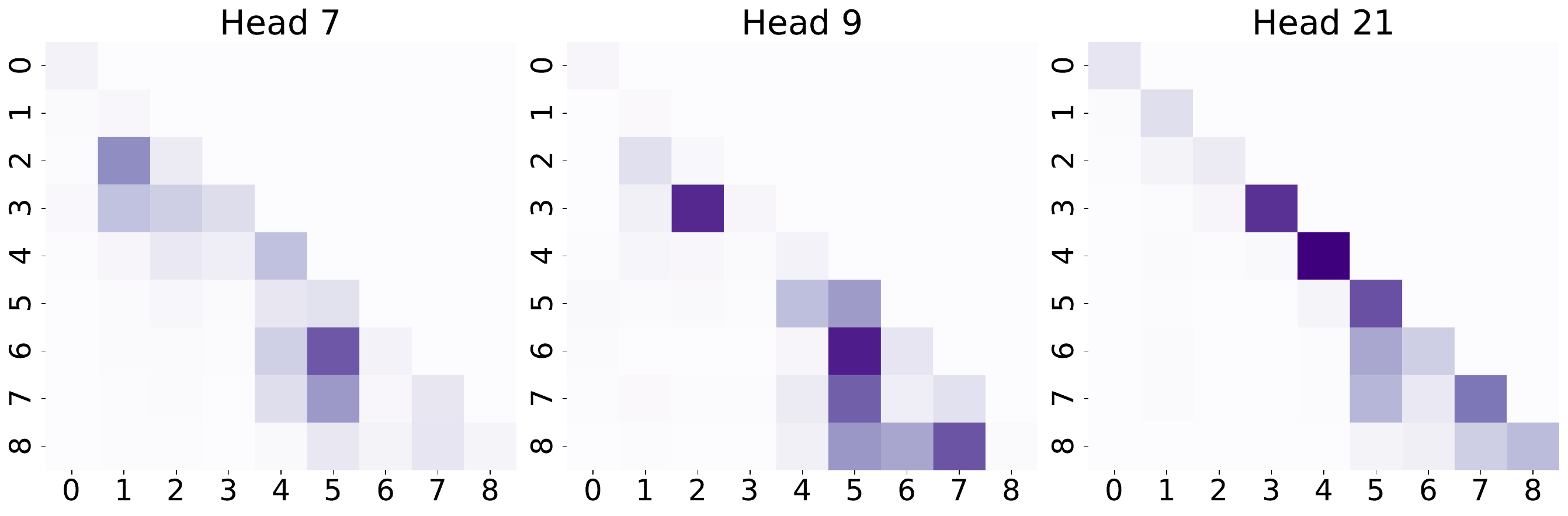}
        \caption{attention scores of 3 different heads for a sample}
        \label{fig:attn-head-multi}
    \end{subfigure}
    \caption{Attention scores from three different heads are visualized for the last 9 tokens of a sample. \fig{fig:attn-head-single} (left) corresponds to the last row of \fig{fig:attn-head-multi} (right), simulating the auto-regressive generation of the final token during inference. Darker colors indicate higher attention scores.}
    \label{fig:attn-head}
    \vspace{-5mm}
\end{figure*}

\citet{dejavu} found that some attention heads show uniform attention scores across previous tokens during the auto-regressive generation. As illustrated in \fig{fig:attn-head-single}, at test time the top head is a uniform “token mixing” head, while the middle and bottom heads are “heavy hitter” heads. Since uniform heads don't capture important interactions, keeping only the heavy hitter heads preserves prediction quality. However, this behavior changes during fine-tuning: \fig{fig:attn-head-multi} shows that attention heads may exhibit different patterns depending on the token—what might be a token-mixing head for one token could be critical for another.


\subsection{Fine-Tuning Details}

\begin{table}[h]
\centering
\setlength{\tabcolsep}{3pt}
\caption{Training Hyperparameters Across Datasets. All experiments use LoRA with dropout = 0, rank = 32, and $\alpha = 64$.}
\begin{tabular}{lcccccc}
\toprule
\textbf{Dataset} & \textbf{Seq. Len} & \textbf{Batch Size} & \textbf{Epochs} & \textbf{LR} & \textbf{Scheduler} & \textbf{Warmup Ratio} \\
\midrule
CSR170K         & 512  & 8  & 1   & 3e-4  & cosine & 0.04 \\
Math10K         & 512  & 8  & 3   & 3e-4  & cosine & 0.04 \\
GLUE (COLA, STS-B, RTE, SST2, QNLI, MNLI, QQP) & 128 & 8  & 3   & 5e-5  & cosine & 0.04 \\
GLUE (MRPC, WNLI)     & 128  & 8  & 5   & 5e-5  & cosine & 0.04 \\
CodeFeedback    & 1024 & 6  & 1  & 2e-5  & cosine & 0.04 \\
WizardLM & 2048 & 2 & 1 & 2e-5  & cosine & 0.04 \\
\bottomrule
\end{tabular}
\label{tab:dataset_details}
\end{table}

\subsection{Detailed Sparsity Configuration}

\begin{table}[h]
\centering
\footnotesize
\setlength{\tabcolsep}{2pt}
\caption{Sparsity Configuration Across Models and Datasets}
\begin{tabular}{lcccccccc}
\toprule
\textbf{Model} & \textbf{Dataset} & \multicolumn{2}{c}{\textbf{FFN}} & \multicolumn{2}{c}{\textbf{QKVO}} & \textbf{Step} & \textbf{\#FLOPs} & \textbf{Speedup} \\
\cmidrule(lr){3-4} \cmidrule(lr){5-6}
 & & \textbf{Layers} & \textbf{Sparsity (\%)} & \textbf{Layers} & \textbf{Sparsity (\%)} & & & \\
\midrule
\multirow{3}{*}{\textbf{LLaMA2-7B}} 
& CSR170K & \multirow{2}{*}{L13--L29} & \multirow{2}{*}{90} & L17--L29 / L20, L24 & 50 & \multirow{2}{*}{5\%} & 0.65 &  $1.3\times$\\
& Math10k & &  & L13--L29 / L20, L24 & 60 &  & 0.73 &  $1.2\times$\\
& CodeFeedback & \multirow{1}{*}{L3--L30}  & \multirow{1}{*}{99} & \multirow{1}{*}{L14--L19,L21--L23,L25-L29}  & \multirow{1}{*}{25} & \multirow{1}{*}{5\%} & 0.73 & $1.2\times$ \\
\midrule
\multirow{2}{*}{\textbf{LLaMA2-13B}} 
& CSR170K & \multirow{2}{*}{L13--L36} & \multirow{2}{*}{97} & \multirow{2}{*}{L17--L36} & 20 & 10\% & 0.61 &  \multirow{2}{*}{$1.3\times$}\\
& Math10k & &  & & 25 & 5\% & 0.70 &  \\
\midrule
\multirow{3}{*}{\textbf{LLaMA3-8B}} 
& CSR170K & L17--L30 & 97 & \multirow{1}{*}{L17--L19,L21--L23,L25-L29} & 20 & \multirow{2}{*}{5\%} & 0.65 & $1.3\times$ \\
& Math10k & L3--L30  & 99 & \multirow{1}{*}{L14--L19,L21--L23,L25-L29}  & 75 &  & 0.46 & $1.6\times$ \\

& GLUE & L17--L30/L31 & 95/50 & L17--L29 & 75 & 5\% & 0.61 & $1.3\times$ \\
\midrule
\multirow{2}{*}{\textbf{LLaMA3.1-8B}}
& CodeFeedback & \multirow{2}{*}{L3--L30}  & \multirow{2}{*}{99} & \multirow{2}{*}{L14--L19,L21--L23,L25-L29}  & \multirow{2}{*}{40} & \multirow{2}{*}{5\%} &  0.66 & $1.3\times$ \\
& WizardLM &   &  &  &  & & 0.53 & $1.5\times$ \\
\bottomrule
\end{tabular}
\end{table}

\subsection{GaLore vs. SparseLoRA}
We compare GaLore~\citep{zhao2024galore} and~\method on CSR170K and Math10K. GaLore training takes slightly more VRAM than LoRA and requires A100 GPUs due to VRAM limitations of A6000 under Distributed Data Parallel (DDP). Runtime is normalized to LoRA, as in our main submission. Result are shown in~\tbl{tab:galore}. GaLore achieves memory efficiency by projecting the full gradient matrices into a low-rank subspace. This projection is periodically updated during training via an online Singular Value Decomposition (SVD) of the gradients. While this online SVD allows GaLore to adapt the subspace and maintain performance similar to LoRA, it incurs a significant $1.58\times$ training overhead compared to LoRA. The amortized time for GaLore, which accounts for these periodic projection updates via online SVD, substantially slows down the fine-tuning process by $13.72\times$. Thus, while GaLore prioritizes memory-efficient fine-tuning by reducing optimizer states and gradient memory, this comes at a considerable cost to computational efficiency due to the demanding SVD operations. In contrast, SparseLoRA is designed to accelerate fine-tuning while delivering near-lossless performance.

\begin{table}[h]
\centering
\small
\begin{tabular}{lcccccccccc}
\toprule
\multicolumn{11}{c}{\textbf{CSR170K}} \\
\midrule
Model & Runtime & Average & BoolQ & PIQA & SIQA & HellaSwag & WinoG & ARC-E & ARC-C & OBQA \\
\midrule
LoRA             & 1.00          & 87.1 & 74.5 & 89.6 & 82.8 & 95.3 & 88.4 & 93.1 & 84.4 & 88.8 \\
GaLore           & 1.58\,[13.72] & 84.1 & 71.2 & 87.1 & 79.6 & 92.0 & 85.0 & 89.4 & 80.5 & 87.8 \\
\method          & 0.78          & 87.0 & 74.7 & 89.5 & 82.8 & 95.3 & 88.8 & 92.9 & 83.6 & 88.3 \\
\bottomrule
\end{tabular}

\vspace{0.8em}

\begin{tabular}{lccccc}
\toprule
\multicolumn{6}{c}{\textbf{Math10K}} \\
\midrule
Model & Runtime & Average & GSM8K & SVAMP & MAWPS \\
\midrule
LoRA             & 1.00          & 80.0 & 71.1 & 79.5 & 89.5 \\
GaLore           & 1.58\,[13.72] & 78.7 & 68.1 & 77.9 & 90.2 \\
\method          & 0.82          & 80.0 & 70.9 & 79.4 & 89.9 \\
\bottomrule
\end{tabular}

\caption{LLaMA 3-8B on CSR170K and Math10K. GaLore’s amortised cost (including periodic online-SVD updates) is shown in brackets.}
\label{tab:galore}
\end{table}

\subsection{Impacts of Learning Rates Sweep}
We perform learning rate sweeps to eliminate selection bias from hyperparameter choices. Specifically, we evaluate both LoRA and SparseLoRA variants using LLaMA3-8B on the Math10K and CSR170K datasets. \tbl{tab:lrsweep_combined}~shows that the performance gap between the best-performing LoRA and \method is just 0.2\% on Math10K and 0.3\% on CSR170K, validating the robustness of our approach.
\begin{table}[h]
\centering
\small
\begin{tabular}{lccc}
\toprule
\textbf{Dataset} & \textbf{Learning Rate} & \textbf{LoRA} & \textbf{\method} \\
\midrule
\multirow{6}{*}{Math10K}
 & $3.0\times10^{-5}$ & 78.3 & 78.8 \\
 & $5.0\times10^{-5}$ & 78.6 & 79.3 \\
 & $9.5\times10^{-5}$ & 79.6 & 79.8 \\
 & $3.0\times10^{-4}$ & 80.0 & \textbf{80.0} \\
 & $5.0\times10^{-4}$ & \textbf{80.2 }& 79.6 \\
 & $9.5\times10^{-4}$ & 78.1 & 77.3 \\
\midrule
\multirow{4}{*}{CSR170K}
 & $3.0\times10^{-5}$ & 85.7 & 85.6 \\
 & $5.0\times10^{-5}$ & 86.7 & 86.5 \\
 & $9.5\times10^{-5}$ & \textbf{87.7} & \textbf{87.4} \\
 & $3.0\times10^{-4}$ & 87.1 & 87.1 \\
\bottomrule
\end{tabular}
\caption{Learning-rate sweep of LLaMA 3-8B on Math10K and CSR170K. Best accuracy per dataset is in bold.}
\label{tab:lrsweep_combined}
\end{table}

\subsection{LoRA on Different Projections}
We primarily apply sparsity to accelerate the main branch of the model, while keeping the LoRA branches dense. To assess generality, we also conduct additional experiments on Math10K, applying LoRA to the Q, K, V, up, and down projections—following the DoRA setup~\citep{dora}—beyond the configurations explored in the main paper. Results in~\tbl{tab:abl_projections} indicate that the benefits of~\method extend beyond just QKVO, demonstrating its broader applicability.

\begin{table}[h]
\centering
\small
\begin{tabular}{lccc}
\toprule
\textbf{Projection Set} & \textbf{LoRA} & \textbf{\method} & $\Delta$ \\
\midrule
QKVO       & 79.9 & \textbf{80.0} & {+0.1} \\
QKVUD      & 80.3 & \textbf{80.9} & {+0.6} \\
QKVOGUD    & 80.5 & \textbf{80.7} & {+0.2} \\
\bottomrule
\end{tabular}
\caption{Mean accuracy of LoRA versus \method on LLaMA 3-8B for different projection configurations.}
\label{tab:abl_projections}
\end{table}

\subsection{Iso-FLOP Comparison}
\begin{figure}[h]
    \centering
    \includegraphics[width=.8\linewidth]{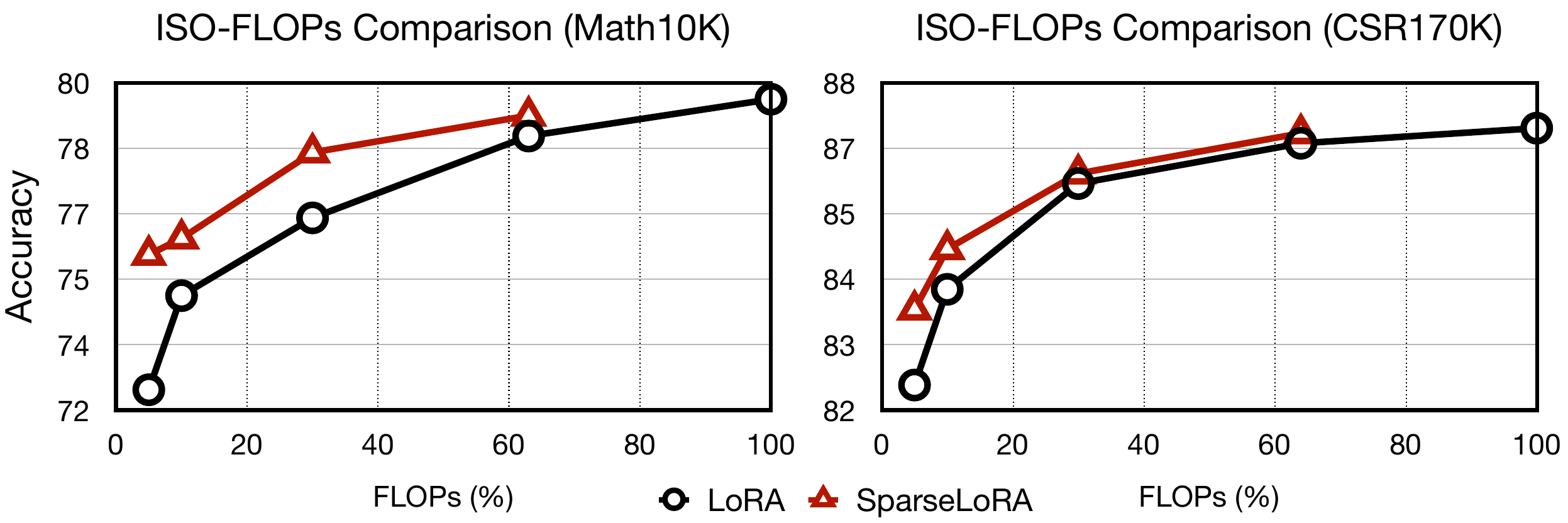}
    \caption{Iso-FLOPs comparison on LLaMA3-8B using Math10K and CSR170K. As the FLOPs budget decreases, \method is able to better retain task performance compared to the LoRA counterpart by a larger margin.}
    \label{fig:iso-flops}
\end{figure}

A practical question is how methods behave when constrained by a \emph{fixed FLOP budget}, rather than a fixed number of training steps. In production, practitioners often allocate a set amount of compute; a method that extracts more accuracy per FLOP is therefore more valuable. To address this, we perform an Iso-FLOP study on LLaMA3-8B using Math10K and CSR170K. One full epoch is treated as 100\% of the available FLOPs, and we sweep down to 5\%.~\fig{fig:iso-flops} shows consistent gains across all tested budgets on both datasets. These results confirm that, for the same computational cost,~\method produces better-performing models than standard LoRA.

\end{document}